\documentclass{article}

\usepackage{microtype}
\usepackage{graphicx}
\usepackage{booktabs} 

\usepackage{hyperref}

\usepackage[accepted]{icml2025}

\usepackage{amsmath}
\usepackage{amssymb}
\usepackage{mathtools}
\usepackage{amsthm}

\usepackage[capitalize,noabbrev]{cleveref}

\theoremstyle{plain}
\newtheorem{theorem}{Theorem}[section]

\newtheorem{corollary}[theorem]{Corollary}
\theoremstyle{definition}
\newtheorem{definition}[theorem]{Definition}

\theoremstyle{remark}

\usepackage[textsize=tiny]{todonotes}

\DeclareMathOperator{\Tr}{Tr}
\DeclareMathOperator{\logdet}{log\,det}

\usepackage{amsfonts}
\usepackage{dsfont}
\usepackage{bm}

\usepackage{caption}
\usepackage{subcaption}

\icmltitlerunning{Learning Kronecker-Structured Graphs from Smooth Signals}

\begin{document}

\twocolumn[
\icmltitle{Learning Kronecker-Structured Graphs from Smooth Signals}



\icmlsetsymbol{equal}{*}

\begin{icmlauthorlist}
\icmlauthor{Changhao Shi}{y1}
\icmlauthor{Gal Mishne}{y2}
\end{icmlauthorlist}

\icmlaffiliation{y1}{Department of Electrical and Computer Engineering, University of California San Diego, the United States}
\icmlaffiliation{y2}{Halıcıoğlu Data Science Institute, University of California San Diego, the United States}

\icmlcorrespondingauthor{Changhao Shi}{cshi@ucsd.edu}
\icmlcorrespondingauthor{Gal Mishne}{gmishne@ucsd.edu}

\icmlkeywords{Machine Learning, ICML}

\vskip 0.3in
]




\begin{abstract}
Graph learning, or network inference, is a prominent problem in graph signal processing (GSP).
GSP generalizes the Fourier transform to non-Euclidean domains, and graph learning is pivotal to applying GSP when these domains are unknown.
With the recent prevalence of multi-way data, there has been growing interest in product graphs that naturally factorize dependencies across different ways.
However, the types of graph products that can be learned are still limited for modeling diverse dependency structures.
In this paper, we study the problem of learning a Kronecker-structured product graph from smooth signals.
Unlike the more commonly used Cartesian product, the Kronecker product models dependencies in a more intricate, non-separable way, but posits harder constraints on the graph learning problem.
To tackle this non-convex problem, we propose an alternating scheme to optimize each factor graph in turn and provide theoretical guarantees for its asymptotic convergence.
The proposed algorithm is also modified to learn factor graphs of the strong product.
We conduct experiments on synthetic and real-world graphs and demonstrate our approach's efficacy and superior performance compared to existing methods.
\end{abstract}

\section{INTRODUCTION}

GSP is a fast-growing field that extends classical signal processing (SP) to non-Euclidean domains \citep{shuman2013emerging,ortega2018graph}.
For a complex system, GSP studies the matrix representation of its graph abstraction.
The spectral decomposition of these graph representations carries important geometric information, from which the Graph Fourier Transform (GFT) is established to analyze and process data that live on the graph.

GSP finds its applications in plenty of fields \citep{borgatti2009network,pavlopoulos2011using,bassett2017network,wu2020comprehensive}, but a prominent problem even before applying GSP is that the graph abstraction of the studied system is frequently unobserved.
Although constructing ad hoc graphs for specific applications may not be difficult, GSP resorts to a more principled method for learning these graphs from the nodal observations (signals).
This data-driven methodology is called graph learning or network topology inference~\citep{mateos2019connecting}.

Graph learning imposes various prior assumptions on the observed data and solves for the graph that fits the best.
Here, we focus on the smoothness assumption, which implies that the observed signals are smooth with respect to the graph of interest.
The smoothness measurement is usually defined as a form of total variation of the graph signals and is related to the combinatorial graph Laplacian.
One can then pose graph learning as an optimization problem regarding the Laplacian matrix.

With the prevalence of multi-way signals or tensors, there has been growing interest in extending GSP to these higher-order structured data.
The graph product then prevails as a convenient tool since the factor graphs naturally capture the mode-wise dependencies of the data \citep{sandryhaila2014big}.
For example, the Cartesian graph product models a non-interactive, parallel composition of factor graphs and serves as the foundation of multi-way GSP \citep{stanley2020multiway}.
However, other types of graph products are still under-explored in GSP.

\begin{figure*}
  \begin{subfigure}[t]{.23\textwidth}
    \centering
    \includegraphics[width=\textwidth]{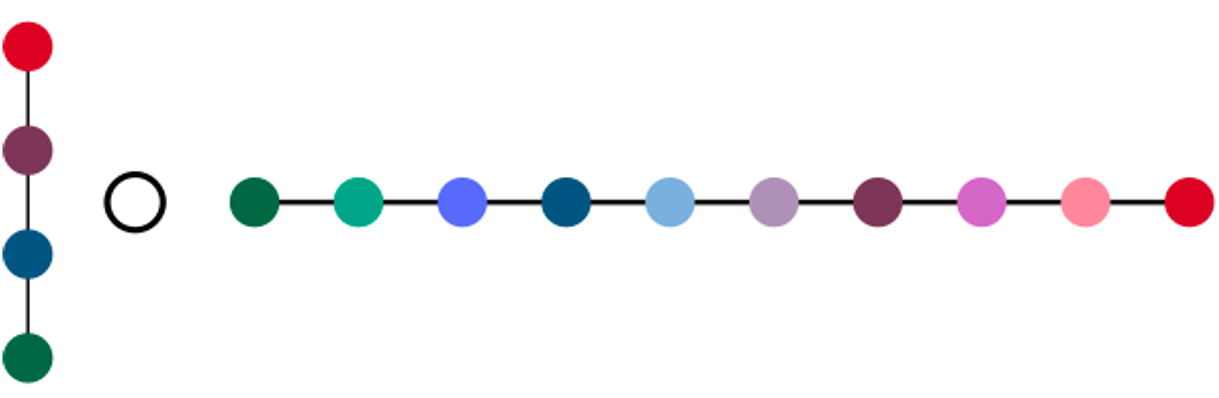}
    \caption{Factor graphs.}
    \label{fig:factors}
  \end{subfigure}
    \hspace{0.01\textwidth}
  \begin{subfigure}[t]{.23\textwidth}
    \centering
    \includegraphics[width=\textwidth]{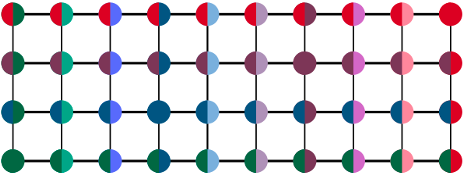}
    \caption{Cartesian product.}
    \label{fig:cart}
  \end{subfigure}
    \hspace{0.01\textwidth}
  \begin{subfigure}[t]{.23\textwidth}
    \centering
    \includegraphics[width=\textwidth]{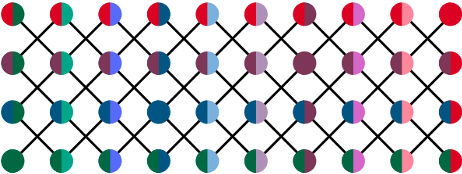}
    \caption{Kronecker product.}
    \label{fig:kron}
  \end{subfigure}
    \hspace{0.01\textwidth}
  \begin{subfigure}[t]{.23\textwidth}
    \centering
    \includegraphics[width=\textwidth]{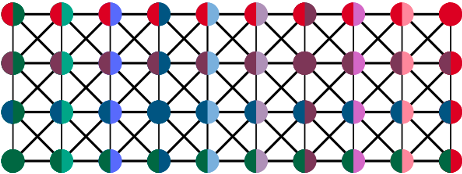}
    \caption{Strong product}
    \label{fig:strong}
  \end{subfigure}
  \caption{An example that compares the Cartesian, Kronecker, and strong graph products.}
  \label{fig:prod_compare}
\end{figure*}

The Kronecker graph product is a powerful model to simulate realistic graphs \citep{leskovec2010kronecker}.
Fig.~\ref{fig:kron} shows an example of the Kronecker graph product.
Unlike the Cartesian product, the Kronecker product wires factor graphs recursively to create a hierarchy with self-similarity.
This model is shown to be useful for mimicking the network characteristics, such as degree distributions of real-world graphs.
These beneficial properties pose the Kronecker product graph as worthy candidates for modeling multi-dimensional structures in GSP \citep{sandryhaila2014big}.
Subsequently, how to learn rigorous Kronecker product graphs from the data emerges naturally as an interesting problem. 

In this paper, we study the problem of learning the Kronecker product graph Laplacian from smooth multi-dimensional data.
GSP has a probabilistic interpretation using the language of graphical models (GM), and graph learning from smooth signals boils down to the parameter estimation of the improper Gaussian Markov random field (IGMRF).
We follow a similar route and formulate our problem as the penalized MLE of an IGMRF with Kronecker product constraints. 
As the problem is not jointly convex, we propose an algorithm that alternates between the optimization of each factor graph.
We also provide theoretical results for the asymptotic convergence of the alternating algorithm, showing an improved convergence rate compared to when the product structure is not accounted for.
Given that the strong graph product also bears a similar Kronecker product form, we also propose a variant of our algorithm to learn strong product graphs from smooth signals.
We conduct experiments on synthetic and real-world graphs and demonstrate our approach's efficacy and superior performance compared to existing methods.
The connections and differences between our method and related GSP and GM methods will also be discussed.
To summarize our contributions:
\begin{itemize}
    \item We are the first to consider the penalized MLE of Kronecker product graph Laplacian learning, and gain theoretical results on its asymptotic consistency, to the best of our knowledge.
    \item We propose a new algorithm to solve the penalized MLE and a variant of it to solve strong product graph Laplacian learning.
    \item We demonstrate that our approach outperforms existing GSP and GM methods on synthetic and real-world datasets.
\end{itemize}

\paragraph{Notations:}
We use the following notations throughout the paper.
Lower-case and upper-case \textbf{bold} letters denote vectors and matrices respectively, and lower-case bold \emph{italic} letters denote random vectors.
Let $\mathbf{1}$ and $\mathbf{0}$ denote the all 1 and all 0 vectors, and let $\mathbf{O}$ denote the all 0 matrix.
Let $\mathbf{e}_p^l\in\mathbb{R}^{p}$ denote a unit vector that has $1$ in its $l$-th entry. 
$\dagger$ denotes the Moore-Penrose pseudo-inverse and ${\det}^\dagger$ denotes the pseudo-determinant. 
$\circ$ denotes the Hadamard product. 
$\otimes$ and $\oplus$ denote the Kronecker product and the Kronecker sum of two matrices, respectively.
$\times$, $\square$, and $\boxtimes$ are used to denote the Kronecker product, the Cartesian product, and the strong product of two graphs, respectively. 
With abuse of notation, $\times$ also denotes the Cartesian product of two sets.
For a node pair $(v,u)$, $\sim$ denotes an edge connects them, and $\not\sim$ denotes non-connection.
For matrix norms, ${\|\cdot\|}_F$ denotes the Frobenius norm, ${\|\cdot\|}_2$ the operator norm,
and ${\|\cdot\|}_{1,\mathrm{off}}$ the sum of the absolute values of all off-diagonal elements.
For random variables, ${\|\cdot\|}_{\psi_2}$ denotes the sub-Gaussian norm.
${(\ \cdot\ )}_+$ denotes the non-negative projection.
$[\ \cdot\ ]_{I,J}$ denotes the sub-matrix of a $n\times m$ matrix at a subset of indices $(I,J)$, where $I \subseteq \{1,2,\dots,n\}$ and $J \subseteq \{1,2,\dots,m\}$.

\section{BACKGROUND}
\label{sec:bg}
\subsection{Graph Representations}
\label{sec:prelim}

Consider an undirected, connected graph $G$ with $|\mathcal{V}|=p$ vertices and $|\mathcal{E}|$ edges.
A graph representation of $G$ is a matrix that fully determines the topology of $G$.
One of the most common graph representations is the weighted symmetric adjacency matrix $\mathbf{W} \in \mathbb{S}^{p}$.
Each entry of the weight matrix ${[\mathbf{W}]}_{ij}={[\mathbf{W}]}_{ji}\geq0$ encodes the weight of a node pair $(i,j)$, and ${[\mathbf{W}]}_{ij}={[\mathbf{W}]}_{ji}>0$ iff $e_{ij}\in \mathcal{E}$.
We assume there are no self-loops, i.e. $\mathbf{W}_{ii}=0$. 
Another important graph representation is the combinatorial graph Laplacian matrix $\bf L$.
The Laplacian of the graph $G$ is defined as $\mathbf{L} =\textbf{D}-\mathbf{W}$, where $\textbf{D}$ denotes the diagonal degree matrix where ${[\mathbf{D}]}_{ii}=\sum_j {[\mathbf{W}]}_{ij}$.
The Laplacian matrix is positive semi-definite by definition, i.e. $\mathbf{L}\in\mathbb{S}_{+}^p$, with the number of zero eigenvalues equal to the number of connected components in the graph.
The Laplacian matrix plays a vital role in spectral graph theory, graph machine learning, and many other scientific fields \citep{merris1994laplacian}.

Let $\mathbf{w}\in\mathbb{R}^{p(p-1)/2}$ denote the vectorization of the graph weights, where ${[\mathbf{w}]}_{i-j+\frac{1}{2}(j-1)(2p-j)}={[\mathbf{W}]}_{ij},\forall 1\leq j\leq i\leq p$.
By definition, $\mathbf{w}$ is also a graph representation.
We then define the linear maps from this non-negative weight vector to its corresponding weighted adjacency matrix and combinatorial graph Laplacian, following \citep{kumar2020unified}.
These linear maps pave the way for our derivation since we will use different graph representations throughout the paper.

\begin{definition}
    Define $\mathcal{A}: \mathbb{R}^{p(p-1)/2}\to\mathbb{R}^{p\times p}, \mathbf{w}\mapsto\mathcal{A}\mathbf{w}$ as the following linear operator
    \[
    {[\mathcal{A}\mathbf{w}]}_{ij} = 
    \begin{cases}
    -{[\mathbf{w}]}_{i-j+\frac{1}{2}(j-1)(2p-j)} & i>j,\\
    {[\mathcal{A}\mathbf{w}]}_{ji} & i<j,\\
    0 & i=j.
    \end{cases}
    \]
\end{definition}
\begin{definition}
    Define $\mathcal{L}: \mathbb{R}^{p(p-1)/2}\to\mathbb{R}^{p\times p}, \mathbf{w}\mapsto\mathcal{L}\mathbf{w}$ as the following linear operator
    \[
    {[\mathcal{L}\mathbf{w}]}_{ij} = 
    \begin{cases}
    -{[\mathbf{w}]}_{i-j+\frac{1}{2}(j-1)(2p-j)} & i>j,\\
    {[\mathcal{L}\mathbf{w}]}_{ji} & i<j,\\
    -\sum_{k\neq j}{[\mathcal{L}\mathbf{w}]}_{kj} & i=j.
    \end{cases}
    \]
\end{definition}
It is obvious that $\mathbf{W}=\mathcal{A}\mathbf{w}$.
One can verify that $\mathcal{L}\mathbf{w}$ is a combinatorial graph Laplacian with weights $\mathbf{w}$.
We then define their adjoint operators.
\begin{definition}
    Define $\mathcal{A}^*: \mathbb{R}^{p\times p}\to\mathbb{R}^{p(p-1)/2}, \mathbf{Q}\mapsto\mathcal{A}^*\mathbf{Q}$ as the following
    \[
    {[\mathcal{A}^*\mathbf{Q}]}_l = \frac{1}{2}({[\mathbf{Q}]}_{ij}+{[\mathbf{Q}]}_{ji}),\
    \]
    \[
    l = i-j+\frac{1}{2}(j-1)(2p-j),\ i>j.
    \]
\end{definition}
\begin{definition}
    Define $\mathcal{L}^*: \mathbb{R}^{p\times p}\to\mathbb{R}^{p(p-1)/2}, \mathbf{Q}\mapsto\mathcal{L}^*\mathbf{Q}$ as the following
    \[
    {[\mathcal{L}^*\mathbf{Q}]}_l = {[\mathbf{Q}]}_{ii}-{[\mathbf{Q}]}_{ij}-{[\mathbf{Q}]}_{ji}+{[\mathbf{Q}]}_{jj},\
    \]
    \[
    l = i-j+\frac{1}{2}(j-1)(2p-j),\ i>j.
    \]
\end{definition}

\subsection{Smoothness Prior in GSP and GM}
\label{sec:sp}

We consider a vector-valued function ${\bm{\mathit{f}}}:\mathcal{V} \rightarrow \mathbb{R}^p$, which assigns a scalar value to each vertex of the graph.
The combinatorial graph Laplacian induces a quadratic $\bm{\mathit{f}}^T\mathbf{L}\bm{\mathit{f}}$, also known as the Dirichlet energy.
The Laplacian quadratic term measures the smoothness (variation) of $\bm{\mathit{f}}$ with respect to $G$, as $\bm{\mathit{f}}^T\mathbf{L}\bm{\mathit{f}}=\sum_{ij}{[\mathbf{W}]}_{ij}{({[\bm{\mathit{f}}]}_i-{[\bm{\mathit{f}}]}_j)}^2$ can be shown.
Given $n$ graph signals $\{\mathbf{f}_1,\mathbf{f}_2,\dots,\mathbf{f}_n\}$,
the inner product $\langle \mathbf{L},\mathbf{S} \rangle=\Tr{(\textbf{LS})}$ of the Laplacian and the sample covariance matrix (SCM) $\mathbf{S}=\frac{1}{n}\sum_{k=1}^n\mathbf{f}_k\mathbf{f}_k^T$ measures the overall smoothness of these signals with respect to the graph.
GSP tackles graph learning by solving the penalized objective function
\begin{equation}
\label{eq:gsp_obj}
    \min_{\mathbf{L} \in \Omega_\mathbf{L}}  \left\{ \langle \mathbf{L}, \mathbf{S} \rangle + \alpha h(\mathbf{L}) \right\},
\end{equation}
where $\Omega_\mathbf{L}$ is the space of all combinatorial graph Laplacians:
\begin{equation}
\label{eq:lp_set}
    \Omega_\mathbf{L} := \left\{ \mathbf{L} \in \mathbf{S}_{+}^p \ | \ \mathbf{L}\mathbf{1}=\mathbf{0}, {[\mathbf{L}]}_{ij}={[\mathbf{L}]}_{ji} \leq 0, \forall i \neq j  \right\},\nonumber
\end{equation}
where $h(\mathbf{L})$ is a penalty term, and $\alpha>0$ is a trade-off parameter.
Minimizing $\langle \mathbf{L}, \mathbf{S} \rangle$ ensures signals $\{\mathbf{f}_k\}$ to vary smoothly on the inferred graph $\mathbf{L}$. 
The penalty term $h(\mathbf{L})$ prevents trivial solutions such as $\mathbf{L}=\mathbf{O}$, and it often encourages other structural properties such as sparsity.

GM approaches the graph learning problem from a different path. 
Consider an IGMRF $\bm{\mathit{f}} \sim \mathcal{N}(\mathbf{0},\mathbf{L}^\dagger)$, 
the penalized MLE is reminiscent of the graphical lasso
\begin{equation}
\label{eq:ggm_mle}
    \min_{\mathbf{L} \in \Omega_\mathbf{L}}  \left\{ \langle \mathbf{L}, \mathbf{S} \rangle - \log{{\det}^\dagger(\mathbf{L})} + \alpha {\|\mathbf{L}\|}_{1,\mathrm{off}} \right\}.
\end{equation}
The additional $\ell_1$ penalization promotes sparsity, and $\alpha>0$ controls its strength.
The Laplacian constraint is what makes \eqref{eq:ggm_mle} substantially different from covariance selection, 
since the solution spaces of these two problems are disjoint. 
Laplacian matrices in $\Omega_\mathbf{L}$ are singular with constant 0-eigenvectors. 
They are also attractive, only allowing positive conditional dependencies.
Covariance selection solves for ordinary precision matrices which are non-singular and have both positive and negative dependencies.
Also notice that one obtains \eqref{eq:gsp_obj} by substituting the penalization in \eqref{eq:ggm_mle} with $-\log{{\det}^\dagger(\mathbf{L})}+\alpha {\|\mathbf{L}\|}_{1,\mathrm{off}}$.
This demonstrates the connection between the GSP and GM formulations.

\section{KRONECKER STRUCTURED GRAPH LEARNING}
\label{sec:pgl}

\subsection{Product Graph Learning}
\label{sec:pl_problem}

Consider two factor graphs $G_1=\{\mathcal{V}_1,\mathcal{E}_1,\mathbf{W}_1\}$ and $G_2=\{\mathcal{V}_2,\mathcal{E}_2,\mathbf{W}_2\}$, with cardinality $|\mathcal{V}_1|=p_1$ and $|\mathcal{V}_2|=p_2$.
A graph product takes $G_1$ and $G_2$ and produces a larger graph $G$ of $|\mathcal{V}| =|\mathcal{V}_1 \times \mathcal{V}_2|=p_1 p_2$ vertices.
Two vertices $(v_1,v_2)$ and $(u_1,u_2)$ in the product graph $G$ are connected iff some product-specific conditions are satisfied.
An example is the Cartesian product $G = G_1 \square G_2$, where $(v_1,v_2) \sim (u_1,u_2)$ holds iff $v_1 = v_2 \wedge u_1 \sim u_2$ or $v_1 \sim v_2 \wedge u_1 = u_2$.
The weighted adjacency matrix of $G$ is $\mathbf{W}_1 \oplus \mathbf{W}_2$.

Although the Cartesian graph product is widely used for modeling non-interactive dependency structures, many applications desire more intricate, interactive structures \citep{sandryhaila2014big}.
As an example, in a time-vertex structure in which nodes `interact' across time, such as the ones in neuroscience, communication, and traffic flows; there exist clear dependencies between neighboring nodes at adjacent time points.
In this setting, the Cartesian product structure is over-simplified and incapable of modeling such dependencies, so we turn to other graph products.

We focus on the Kronecker and strong products, which are two other common options for modeling product graphs.
The Kronecker product of $G_1$ and $G_2$ is denoted as $G = G_1 \times G_2$, where $(v_1,v_2) \sim (u_1,u_2)$ iff $v_1 \sim v_2 \wedge u_1 \sim u_2$.
The weighted adjacency matrix of $G$ is the Kronecker product of the factor weights $\mathbf{W}_1 \otimes \mathbf{W}_2$.
Another graph product that produces even denser connectivity is the strong product.
The strong product $G = G_1 \boxtimes G_2$ is defined as the union of the Kronecker and the Cartesian products.
Thus the weighted adjacency matrix of the strong product graph is $\mathbf{W}_1 \otimes \mathbf{W}_2 + \mathbf{W}_1 \oplus \mathbf{W}_2$.
Fig.~\ref{fig:prod_compare} illustrates the aforementioned three graph products.

We now formulate the product graph learning problem using the Kronecker graph product.
This formulation also enlightens strong product graph learning, as discussed later.
Let the random matrix $\bm{\mathit{X}}\in\mathbb{R}^{p_1\times p_2}$ represent a two-way graph signal that lives on the product graph $G$. 
${[\bm{\mathit{X}}]}_{i_1,i_2}$ is the signal on node $(i_1,i_2)$.
Given $n$ instantiations $\{\textbf{X}_1,\textbf{X}_2,\dots,\textbf{X}_n\}$, our goal is to learn the factor graphs $G_1, G_2$ and their Kronecker product $G$ from these nodal observations on $G$.
Note that our argument can be generalized to more factors naturally, though not presented here.

Let the random vector $\bm{\mathit{x}}$ be the vectorization of $\bm{\mathit{X}}$ and $\mathbf{S}=\frac{1}{n}\sum_{k=1}^{n}\textbf{x}_k{\textbf{x}_k}^T$ be the SCM.
Since for $G = G_1 \times G_2$ we have $\mathbf{A}=\mathbf{A}_1\otimes\mathbf{A}_2$, we derive the non-penalized product graph learning objective 
\begin{equation}
\label{eq:pggm_mle}
\begin{gathered}
    \min_{\mathbf{w}_1,\mathbf{w}_2 \in \Omega_\mathbf{w}}  \Bigl\{ \langle \mathbf{L}, \mathbf{S} \rangle - \log{{\det}^\dagger(\mathbf{L})}\Bigr\}, \\
    \textrm{s.t. }
    \mathbf{L} = \mathbf{D}_1 \otimes \mathbf{D}_2 - \mathbf{W}_1 \otimes \mathbf{W}_2
\end{gathered}
\end{equation}

It is worth to emphasize that unlike for Cartesian product graphs, $\mathbf{L}\neq\mathbf{L}_1\otimes\mathbf{L}_2$ and thus $\log{{\det}^\dagger(\mathbf{L})} \neq \log{{\det}^\dagger(\mathbf{L}_1)} + \log{{\det}^\dagger(\mathbf{L}_2)}$.
This makes our problem substantially different from the MLE of matrix normal distributions \citep{dutilleul1999mle}.
Interestingly, except for the GSP merits, the Laplacian constraints also endow total positivity \citep{lauritzen2019maximum}, a GM property that is not compatible with other Kronecker structured distributions.

\subsection{Kronecker Product Graphs} 
\label{sec:KSGL}

We propose the \textbf{K}ronecker \textbf{S}ructured \textbf{G}raph (Laplacian) \textbf{L}earning (\textbf{KSGL}) algorithm for solving \eqref{eq:pggm_mle}.
We formulate the penalized MLE \eqref{eq:pggm_mle} as
\begin{align}
\label{eq:mle_pgd}
        \min_{\mathbf{w}_1,\mathbf{w}_2 \geq \mathbf{0}}  \Bigl\{ & \langle \mathcal{A}\mathbf{w}_1\otimes\mathcal{A}\mathbf{w}_2, \mathbf{K} \rangle + \alpha_1\mathbf{w}_1^T\mathbf{1} + \alpha_2\mathbf{w}_2^T\mathbf{1} - \nonumber \\ & \log{{\det}^\dagger(\mathcal{L}\mathcal{A}^*(\mathcal{A}\mathbf{w}_1\otimes\mathcal{A}\mathbf{w}_{2}))}  \Bigr\},
\end{align}
with $\ell_1$ penalization.
Here $\mathbf{K}=\mathcal{A}\mathcal{L}^*\mathbf{S}$ denotes the pairwise square 
Euclidean distances of the signals.
The absolute sign of the $\ell_1$ norm of the sparsity penalty is redundant due to the non-negative constraints.
KSGL then operates in an alternating scheme to solve $\mathbf{w}_1$ and $\mathbf{w}_2$.
The algorithm starts with initialization $\mathbf{w}_1=\frac{1}{p_1}\mathbf{1}_{\frac{p_1(p_1-1)}{2}}$ and $\mathbf{w}_2=\frac{1}{p_2}\mathbf{1}_{\frac{p_2(p_2-1)}{2}}$.
It then uses projected gradient descent to solve for one variable while keeping the other fixed until the stopping criteria are met.
The update of $\mathbf{w}_1$ and $\mathbf{w}_2$ is given by
\begin{align}
    {[\mathbf{w}_1^{(t+1)}]}_{m_1} = & \Bigl( {[\mathbf{w}_1^{(t)}]}_{m_1}-\eta( \langle \mathbf{W}_2^{(t)}, {[\mathbf{K}]}_{I_1,J_1} \rangle - \label{eq:update_w1}\\
    & \langle \mathbf{W}_2^{(t)}, {[\mathcal{A}\mathcal{L}^*{(\mathbf{L}^{(t)}+\mathbf{J})}^{-1}]}_{I_1,J_1} \rangle + \alpha_1) \Bigr)_+\nonumber,
    \end{align}
    \begin{align}
    {[\mathbf{w}_2^{(t+1)}]}_{m_2} = & \Bigl( {[\mathbf{w}_2^{(t)}]}_{m_2}-\eta( \langle \mathbf{W}_1^{(t+1)}, {[\mathbf{K}]}_{I_2,J_2} \rangle - \label{eq:update_w2}\\
    & \langle \mathbf{W}_1^{(t+1)}, {[\mathcal{A}\mathcal{L}^*{(\mathbf{L}^{(t')}+\mathbf{J})}^{-1}]}_{I_2,J_2} \rangle + \alpha_2) \Bigr)_+\nonumber.
\end{align}
Here $m_1=i-j+\frac{1}{2}(j-1)(2p_1-j)$ and $m_2=i-j+\frac{1}{2}(j-1)(2p_2-j)$.
The subsets $I_1=\{(i-1)p_2+1, (i-1)p_2+2, \dots, ip_2\}$ and $J_1=\{(j-1)p_2+1, (j-1)p_2+2, \dots, jp_2\}$ specify node pairs associated with ${[\mathbf{w}_1]}_{m_1}$, and similarly for the subsets $I_2=\{i, p_2+i, \dots, (p_1-1)p_2+i\}$ and $J_2=\{j, p_2+j, \dots, (p_1-1)p_2+j\}$.
Alg.~\ref{alg:kpgl} summarizes the algorithm.

\subsection{Strong Product Graphs} 
\label{sec:ssgl}

An alternative graph product with broad applications is the strong graph product.
We now demonstrate how the strong product relates to the Kronecker product and how we can easily modify KSGL to learn strong product graphs.
For factor graphs $G_1$ and $G_2$, consider adding self-loops to them and then taking the Kronecker product.
The new product graph is also self-looped and its weighted adjacency matrix is $(\mathbf{W}_1+\mathbf{I}_{p_1}) \otimes (\mathbf{W}_2+\mathbf{I}_{p_2}) = \mathbf{W}_1\otimes\mathbf{W}_2 + \mathbf{W}_1\otimes\mathbf{I}_{p_2} + \mathbf{I}_{p_1}\otimes\mathbf{W}_2 + \mathbf{I}_p = (\mathbf{W}_1\otimes\mathbf{W}_2 + \mathbf{W}_1\oplus\mathbf{W}_2) + \mathbf{I}_p$.
Removing the self-loops, we obtain exactly the strong product of $G_1$ and $G_2$.
This relation helps us formulate the penalized MLE for learning strong product graphs based on \eqref{eq:mle_pgd}
\begin{align}
\label{eq:mle_pgd_st}
        \min_{\mathbf{w}_1,\mathbf{w}_2 \geq \mathbf{0}}  \Bigl\{ & \langle (\mathcal{A}\mathbf{w}_1+\mathbf{I}_{p_1})\otimes(\mathcal{A}\mathbf{w}_2+\mathbf{I}_{p_2}), \mathbf{K} \rangle - \nonumber\\& \log{{\det}^\dagger(\mathcal{L}\mathcal{A}^*((\mathcal{A}\mathbf{w}_1+\mathbf{I}_{p_1})\otimes(\mathcal{A}\mathbf{w}_2+\mathbf{I}_{p_2})))} + \nonumber \\ & \alpha_1\mathbf{w}_1^T\mathbf{1} + \alpha_2\mathbf{w}_2^T\mathbf{1} \Bigr\}.
\end{align}
Here we plug in the self-looped strong product adjacency matrix since the pairwise distances are all 0 on the $\mathbf{K}$ diagonal and $\mathcal{A}^*$ is also agnostic to its input diagonal values.
Similarly, we use projected gradient descent to solve for $\mathbf{w}_1$ or $\mathbf{w}_2$ and then alternate between these two steps.
The update of $\mathbf{w}_1$ and $\mathbf{w}_2$ is 
\begin{align}
    \label{eq:update_w1_st} 
  &  {[\mathbf{w}_1^{(t+1)}]}_{m_1} = \Bigl( {[\mathbf{w}_1^{(t)}]}_{m_1}-\eta( \langle \mathbf{W}^{(t)}_2+\mathbf{I}_{p_2}, {[\mathbf{K}]}_{I_1,J_1} \rangle  -  \nonumber \\ 
   &  \langle \mathbf{W}^{(t)}_2+\mathbf{I}_{p_2}, {[\mathcal{A}\mathcal{L}^*{(\mathbf{L}^{(t)}+\mathbf{J})}^{-1}]}_{I_1,J_1} \rangle + \alpha_1) \Bigr)_+, 
     \end{align}
\begin{align}
\label{eq:update_w2_st}
&{[\mathbf{w}_2^{(t+1)}]}_{m_2} = \Bigl( {[\mathbf{w}_2^{(t)}]}_{m_2}-\eta( \langle \mathbf{W}_1^{(t+1)}+\mathbf{I}_{p_1}, {[\mathbf{K}]}_{I_2,J_2} \rangle - \nonumber \\
 &    \langle \mathbf{W}_1^{(t+1)}+\mathbf{I}_{p_1}, {[\mathcal{A}\mathcal{L}^*{(\mathbf{L}^{(t')}+\mathbf{J})}^{-1}]}_{I_2,J_2} \rangle + \alpha_2) \Bigr)_+ .
\end{align}

\begin{algorithm}[tb]
    \caption{KSGL}
    \label{alg:kpgl}
\begin{algorithmic}
    \STATE {\bfseries Input:} graph signals $\{\mathbf{X}_k\}$, parameters $\alpha,\eta$
    \STATE {\bfseries Output:} factor graph weights $\mathbf{w}_1,\mathbf{w}_2$
    \STATE Compute $\mathbf{S}$ and $\mathbf{K}$.
    \STATE Initialize $\mathbf{w}_1$ and $\mathbf{w}_2$.
    \REPEAT
        \REPEAT
            \STATE Update $\mathbf{w}_1$ as in: 
            \STATE $\ \ $ \eqref{eq:update_w1} for the Kronecker product or 
            \STATE $\ \ $ \eqref{eq:update_w1_st} for the strong product
        \UNTIL{convergence.}
        \REPEAT
            \STATE Update $\mathbf{w}_2$ as in: 
            \STATE $\ \ $ \eqref{eq:update_w2} for the Kronecker product or 
            \STATE $\ \ $ \eqref{eq:update_w2_st} for the strong product
        \UNTIL{$\mathbf{w}_2$ convergence.}
    \UNTIL{$\mathbf{w}_1$ and $\mathbf{w}_2$ converge or maximum iterations.}
\end{algorithmic}
\end{algorithm}

\section{THEORETICAL RESULTS}

Here we establish the statistical consistency and convergence rates for the penalized Kronecker product graph Laplacian estimator as in \eqref{eq:mle_pgd}.
We first make assumptions regarding the true underlying graph we were to estimate:
\begin{itemize}
    \item[(A1)] Let $\mathcal{S}_1$ and $\mathcal{S}_2$ be the support set of the true factor graphs. We assume the graphs are sparse and the cardinality of their supports is upper bounded by $|\mathcal{S}_1| \leq s_1p_1$ and $|\mathcal{S}_2| \leq s_2p_2$.
    \item[(A2)] Let $(d_{1,\mathrm{min}}, d_{1,\mathrm{max}})$ and $(d_{2,\mathrm{min}}, d_{2,\mathrm{max}})$ be the minimum and maximum degrees of the true factor graphs. We assume these degrees are bounded away from $0$ and $\infty$ by a constant $d>1$, such that $\frac{1}{d} \leq d_{1,\mathrm{min}}\leq d_{1,\mathrm{max}}\leq d$ and $\frac{1}{d} \leq d_{2,\mathrm{min}}\leq d_{2,\mathrm{max}}\leq d$.
    \item[(A3)] Let $\{0,\lambda_2,\dots,\lambda_p\}$ be the eigenvalues of the true product graph Laplacian in a non-decreasing order.
    We assume these eigenvalues are bounded away from $0$ and $\infty$ by a constant $z>1$, such that $\frac{1}{z} \leq\lambda_2<\lambda_p\leq z$.
\end{itemize}
These assumptions are common in high-dimensional statistics.
They also imply that the product and factor graphs are connected graphs.
With the above assumptions, our first theorem states that a solution to the MLE problem always exists.
Proofs of all the theorems can be found in the supplement.
\begin{theorem}[Existence of MLE]
\label{th:existence}
    The penalized negative log-likelihood of Kronecker product graph Laplacian learning as in \eqref{eq:mle_pgd} is lower-bounded, and there exists at least one global minimizer as the solution of the penalized MLE.
\end{theorem}
Our proof largely follows \citep{ying2021minimax,shi2023graph}, which shows that the objective function is lower-bounded and the minimizer is achievable.
However, note that since the original problem is not jointly convex, the solution is not unique.
In fact, a set of solution $(\mathbf{w}_1^*, \mathbf{w}_2^*)$ is not identifiable to the Kronecker graph product $\mathcal{A}\mathbf{w}_1\otimes\mathcal{A}\mathbf{w}_2$ since $\forall a>0, a \mathbf{w}_1^*, \frac{1}{a} \mathbf{w}_2^*$ is also a solution.
Nevertheless, our Theorem~\ref{th:uniqueness} states that the alternating optimization enjoys a unique solution in each sub-problem.

\begin{theorem}[Uniqueness of MLE]
\label{th:uniqueness}
    The objective function of the penalized MLE is bi-convex with respect to each factor graph, and a global minimizer for each sub-problem uniquely exists.
\end{theorem}
The proof shows that when one of the factors is held fixed, optimizing the other factor becomes a convex problem.
We then show that the Kronecker product graph Laplacian learned by KSGL is asymptotically consistent. 

\begin{corollary}
\label{cl:consistency}
    Suppose assumptions (A1)-(A3) hold for the true factor graphs.
    Then with sufficiently large $n$
    and proper penalty $\alpha_1$ and $\alpha_2$,
    the Frobenius errors of the minimizers $\widehat{\mathbf{L}}_1=\mathcal{L}(\mathbf{\widehat{w}_1})$ and $\widehat{\mathbf{L}}_2=\mathcal{L}(\mathbf{\widehat{w}_2})$ are bounded by
    \begin{align}
    \label{ieq:convergence_rate}
        &{\|\widehat{\mathbf{L}}_1-\mathbf{L}_1^*\|}_F \leq c_1 \sqrt{\frac{s_1p_1\log p}{np_2}},\\
        &{\|\widehat{\mathbf{L}}_2-\mathbf{L}_2^*\|}_F \leq c_2 \sqrt{\frac{s_2p_2\log p}{np_1}}.
    \end{align}
     with high probability.
\end{corollary}
Corollary~\ref{cl:consistency} proves that the solution of a sub-problem converges to the ground truth when the other factor is bounded.
This helps use induction to prove the consistency of \eqref{eq:mle_pgd}.
\begin{theorem}[High-dimensional consistency]
\label{th:consistency}
    Suppose assumptions (A1)-(A3) hold for the true factor graphs.
    Then with sufficiently large $n$,
    proper penalty $\alpha_1$ and $\alpha_2$, and Corollary~\ref{cl:consistency}
    the minimizer $\widehat{\mathbf{L}}=\mathcal{L}(\mathbf{\widehat{w}})$ of the penalized MLE as in \eqref{eq:mle_pgd} is asymptotically consistent to the true Laplacian $\mathbf{L}^*$,
    and the Frobenius error is bounded by
    \begin{equation}
    \label{ieq:convergence_rate}
        {\|\widehat{\mathbf{L}}-\mathbf{L}^*\|}_F \leq c \sqrt{\frac{(p_1+p_2)\log p}{n}}.
    \end{equation}
     with high probability.
\end{theorem}
Theorem~\ref{th:consistency} shows that the learned product graph Laplacian converges to the true Laplacian asymptotically under mild conditions.
The final error depends on the norms of initializations.
Compared with the IGMRF convergence rate from \citep{ying2021minimax}, KSGL converges faster by a factor of $\sqrt{\frac{p_1 p_2}{p_1+p_2}}$ with similar probability.
This shows how leveraging the product structure prior benefits graph learning.
The improvement of the convergence rate is similar to the ones in \citep{tsiligkaridis2013convergence}.

\section{Related Work}
\paragraph{Smooth Graph Learning}
Learning graph Laplacian matrices from smooth signals has been studied extensively in GSP \citep{dong2016learning,kalofolias2016learn,chepuri2017learning,egilmez2017graph,zhao2019optimization,buciulea2022learning}.
These papers focus on the Laplacian quadratic terms, which correspond to the Dirichlet energy of the signals.
\citet{thanou2017learning} model the smoothness differently, using the heat diffusion process. 
\citet{pasdeloup2017characterization} propose to learn the normalized graph Laplacian matrix instead of the combinatorial Laplacian.
The weighted adjacency matrix has also been widely studied as a different graph representation \citep{segarra2017network,shafipour2021identifying}.

\paragraph{Product Graph Learning}
Learning product graphs amounts to posing structural constraints on graph learning.
Previous work mainly focuses on Cartesian product graphs \cite{lodhi2020learning,Kadambari2020learning,kadambari2021product,einizade2023learning,shi2024learning}; few have studied other products such as the Kronecker \citep{lodhi2020learning,einizade2023learning}.
\citet{lodhi2020learning} proposed to learn the factor graphs under the trace constraints;
\citet{einizade2023learning} posited that an accurate eigenbasis estimation of the factor graph shift operator (GSO) is known and solved for the eigenvalues.
However, these methods either do not learn the combinatorial graph Laplacian, rely on different assumptions, or fall short of their theoretical properties.

\paragraph{Matrix Variate Distributions}
It is well-known that the generative process of smooth graph signals can be modeled as a GMRF with a Laplacian precision matrix \citep{zhang2015graph}.
Therefore, methods for covariance selection \citep{dempster1972covariance,banerjee2006convex,yuan2007model,banerjee2008model} that aim to learn sparse precision matrices, such as the graphical lasso \citep{friedman2008sparse}, often serve as additional baselines of graph learning methods.
However, these methods do not learn a rigorous combinatorial graph Laplacian.
The matrix variate normal distribution \citep{dawid1981some,gupta1999matrix} can be seen as a generalization of the GMRF to multi-way signals.
The covariance matrices, and thus the precision matrices, are endowed with a Kronecker product structure, and the graphical lasso algorithm has been extended to learn these Kronecker graphical models \citep{dutilleul1999mle,werner2008estimation,zhang2010learning,leng2012sparse,tsiligkaridis2013convergence}.
Other matrix variate distributions replace the Kronecker product structure with the Kronecker sum \citep{kalaitzis2013bigraphical,greenewald2019tensor,wang2020sylvester,yoon2022eiglasso}, leading to Cartesian product graphs.
While these graphical lasso methods bear a similar form to the Kronecker product graph Laplacian learning, none of these learn precision matrices are Laplacian, thus not appropriate for use in GSP.

\section{EXPERIMENTS}
\label{sec:exp}

We conduct extensive experiments on both synthetic and real-world datasets to evaluate our method.

\subsection{Synthetic Graphs}
\label{sec:exp_synthetic}

\begin{figure*}[hbt!]
\centering
\includegraphics[width=\textwidth]{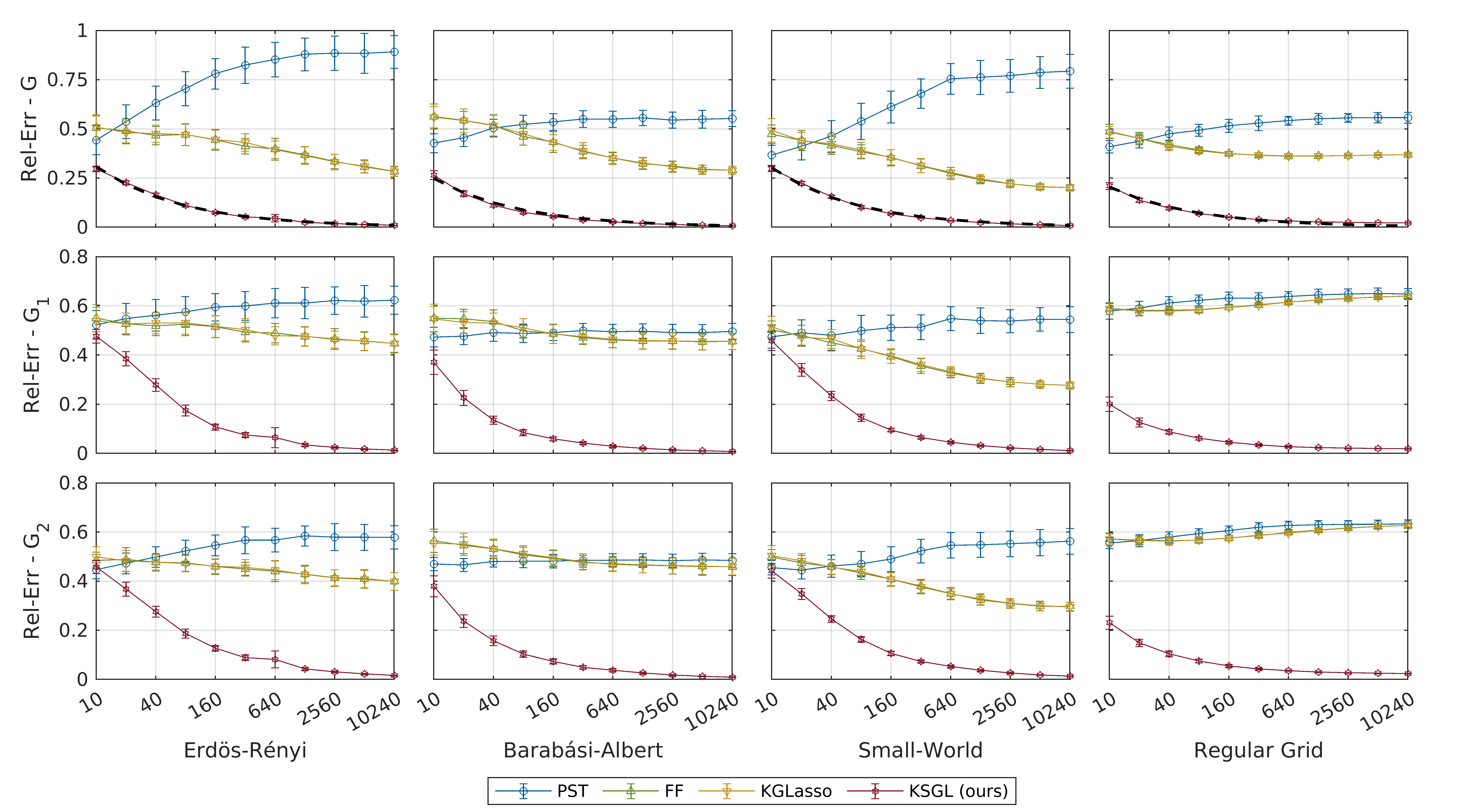}
\caption{Comparison of different methods on various synthetic Kronecker product graphs and signals.
Each sub-figure shows the trend of Rel-Err of the product (top row) or factor (middle and bottom rows) Laplacian matrices as $n$ increases.
Black dash lines fit the theory in \eqref{ieq:convergence_rate} to the KSGL results.}
\label{fig:re}
\end{figure*}

Since the ground truth graphs are often unavailable or not defined in real-world problems, we first evaluate our methods on synthetic signals where the underlying graph to be estimated is known.
We follow \citep{shi2024learning} to generate factor graphs using the graph models below
\begin{enumerate}
    \item[(1)] Erd\H{o}s-R\'enyi model with probability $p=0.3$;
    \item[(2)] Barab\'asi-Albert model with preferential attachment $m=2$ and $m_0=2$ initial nodes;
    \item[(3)] Watts-Strogatz small-world model, where a chain graph with node degree $d=2$ is rewired with probability $p=0.1$;
    \item[(4)] and regular grids.
\end{enumerate}
We set the number of nodes to $p_1=20$ and $p_2=25$ for each factor, and the dimensions of regular grids are $4\times5$ and $5\times5$.
To obtain weighted graphs, we randomly sample a weight from a uniform distribution $\mathcal{U}(0.1,2)$ for each edge.
We then generate the signals from the IGMRF process $\bm{\mathit{f}}\sim \mathcal{N}(\mathbf{0},\mathbf{L}^{\dagger})$, where $\mathbf{L}$ is the Laplacian of the Kronecker product graph or the strong product graph.
The goal of graph learning is to recover the underlying weighted graphs from the signals, where we vary the number of the signals $n=10\times2^r,r\in\{0,1,\dots,10\}$.

We create 50 realizations for each graph and dataset size and report the mean and standard deviation of the selected metrics: the relative error (Rel-Err) of the Laplacian and the area under the precision-recall curve (PR-AUC) of edge prediction.
The former Rel-Err computes the relative Frobenius error of the learned factor and product graph Laplacian matrices to their ground truth counterparts.
To eliminate the ambiguity of the learned factor graphs, we normalize the graph Laplacian matrices by their cardinality $\frac{p\mathbf{L}}{\Tr{(\mathbf{L})}}$ before computing the relative error. 
The latter PR-AUC considers the binary prediction of the ground truth edge patterns.
We choose PR-AUC over ROC-AUC since the two classes are highly imbalanced (edge versus no edge).

We evaluate KSGL against three competing methods that model the Kronecker structures: the PST (Product Spectral Template) method \citep{einizade2023learning}, the FF (Flip-Flop) method \citep{dutilleul1999mle}, and the Kronecker Graphical Lasso (KGLasso) method \citep{tsiligkaridis2013convergence}.
PST is a GSP method that extracts the eigenvectors of factor GSOs from the signal covariance.
FF solves the MLE of matrix normal distributions and KGLasso adds sparsity constraints to that, both of which fall into the GM category.
For the strong product experiments, since the strong graph product is the union of the Kronecker and Cartesian graph product, we add Cartesian product graph learning methods for comparison: the PGL (Product Graph Learning) method \citep{kadambari2021product}, the TeraLasso (Tensor Graphical Lasso) method \citep{greenewald2019tensor}, and the MWGL (Multi-Way Graph Learning) method \citep{shi2023graph}.
We follow the common grid-search procedure in each setting to select the best-performing hyper-parameters for each method.

\begin{figure*}[t!]
\centering
\includegraphics[width=\textwidth]{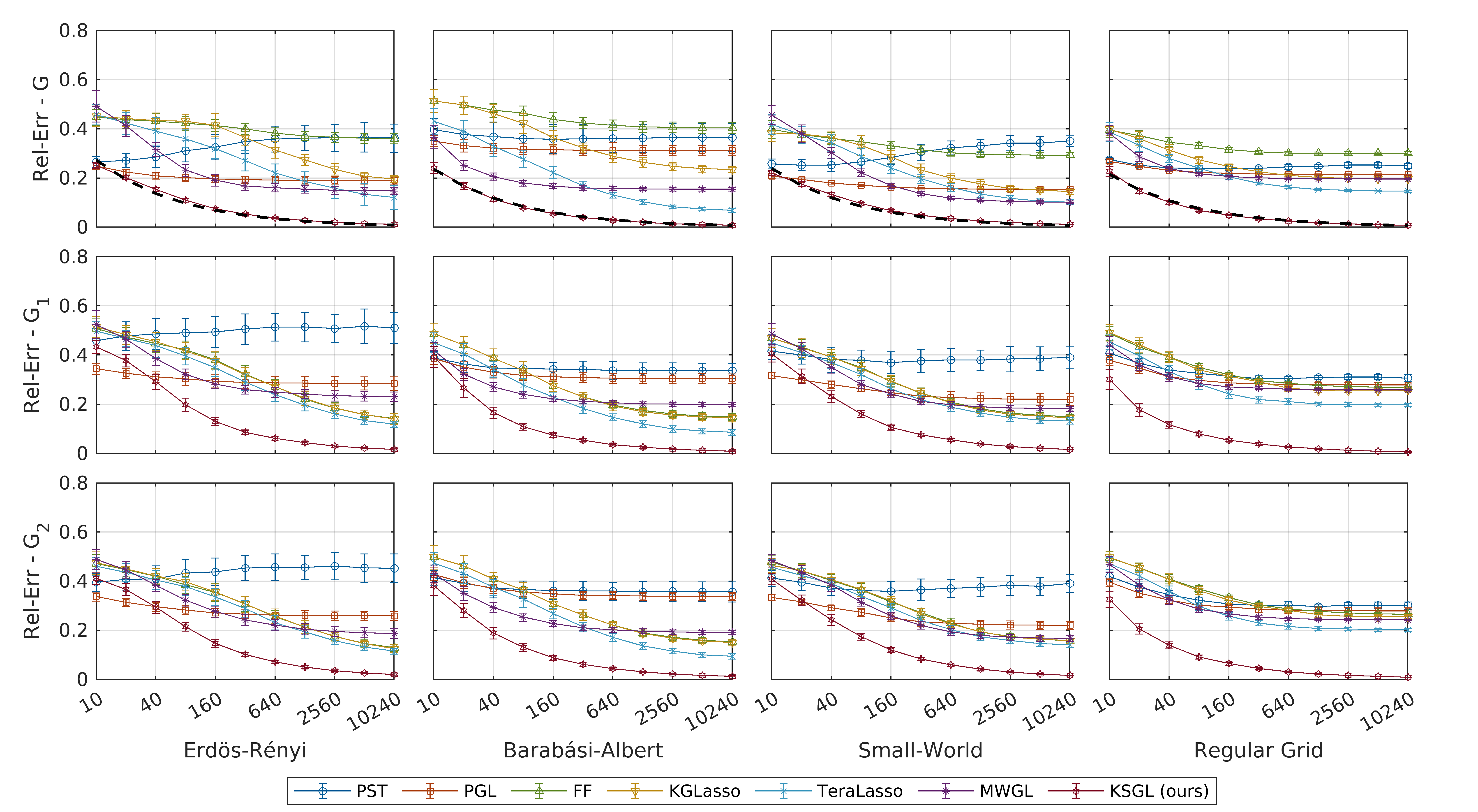}
\caption{Comparison of different methods on various synthetic strong product graphs and signals.
Each sub-figure shows the trend of Rel-Err as $n$ increases.
Black dash lines fit the theory in \eqref{ieq:convergence_rate} to our results.}
\label{fig:re_strong}
\end{figure*}

Fig.~\ref{fig:re} shows the trend of Rel-Err as the number of signals increases in different settings.
We provide the PR-AUC results in Appendix~\ref{sec:pr-auc}. 
As we can see, KSGL outperforms the competing methods in every setting.
The Rel-Err of KSGL converges to 0 as the number of signals increases and its trend validates our theoretical results (black dash lines, top row).
PST does not perform well because the spectral templates cannot be estimated accurately but only roughly approximated even with a large number of signals (see more details in Appendix~\ref{sec:baselines}).
FF and KGLasso also underperform KSGL because of the inherent model mismatch, which is that the Laplacian of a Kronecker product graph is not the Kronecker product of the factor Laplacians.
Another reason is that the precision matrices learned by FF and KGLasso are not Laplacian.
Their similar performance also shows that adding sparsity constraints to the wrong assumption does not benefit Kronecker product graph learning.

Fig.~\ref{fig:re_strong} shows the Rel-Err results of strong product graph learning.
Again KSGL behaves advantageously in almost every setting.
The only exception is that PGL performs better in low data regimes on Erd\H{o}s-R\'enyi and Watts-Strogatz small-world graphs, but it fails to deliver as the number of signals increases due to the model mismatch.
Among other competing methods, TeraLasso and MWGL perform the best overall, but they still fall behind KSGL by a margin.
Note that we do not include PGL, TeraLasso, and MWGL in the Kronecker product experiments since these Cartesian product methods are expected to fail in these settings.
We verify this using Erd\H{o}s-R\'enyi graphs in Appendix~\ref{sec:pr-auc}.

\subsection{EEG Data}
\label{sec:exp_eeg}

We now evaluate KSGL on real EEG recordings \citep{nasreddine2021eeg}.
The EEG data are collected from epileptic patients using the 10-20 electrode system.
The signals from 21 scalp electrodes are divided into 1-second segments, and we sub-sample the signals to get 50 samples per segment.
Each segment is also annotated with a class label, indicating if the patient is undergoing a seizure and further which type of seizure it is.
This results in 21$\times$50 multi-way signals of different categories from multiple patients. 
Our goal is to learn a graph of brain region (electrodes) and a graph of time from these multi-way signals.

\begin{figure}[hbt!]
    \centering
    \begin{subfigure}[b]{0.2\textwidth}
        \centering
        \includegraphics[width=\textwidth]{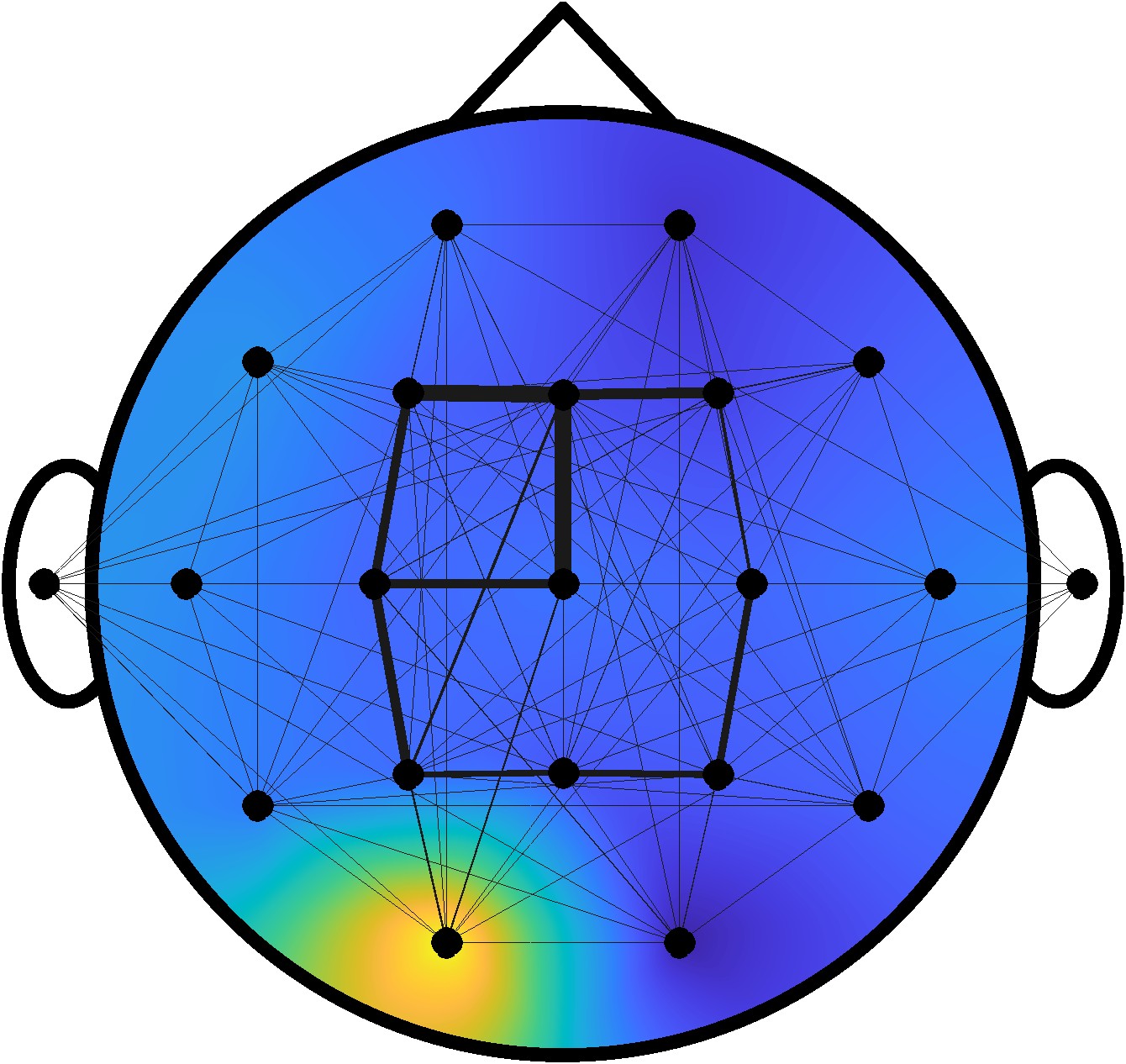}
        \caption{MWGL - normal}
    \end{subfigure}
    \hspace{0.01\textwidth}
    \begin{subfigure}[b]{0.2\textwidth}
        \centering
        \includegraphics[width=\textwidth]{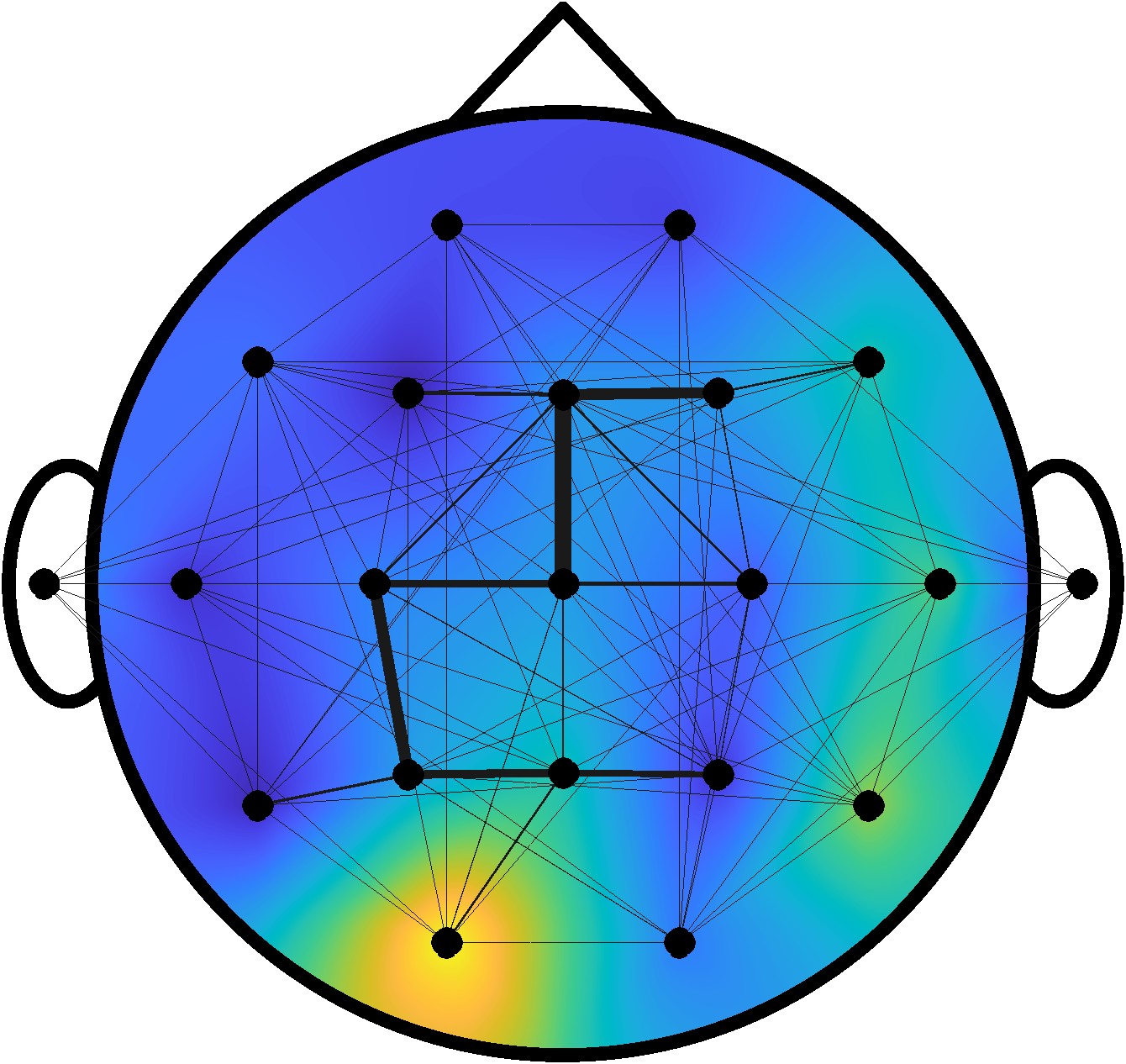}
        \caption{MWGL - seizure}
    \end{subfigure}
    \begin{subfigure}[b]{0.2\textwidth}
        \centering
        \includegraphics[width=\textwidth]{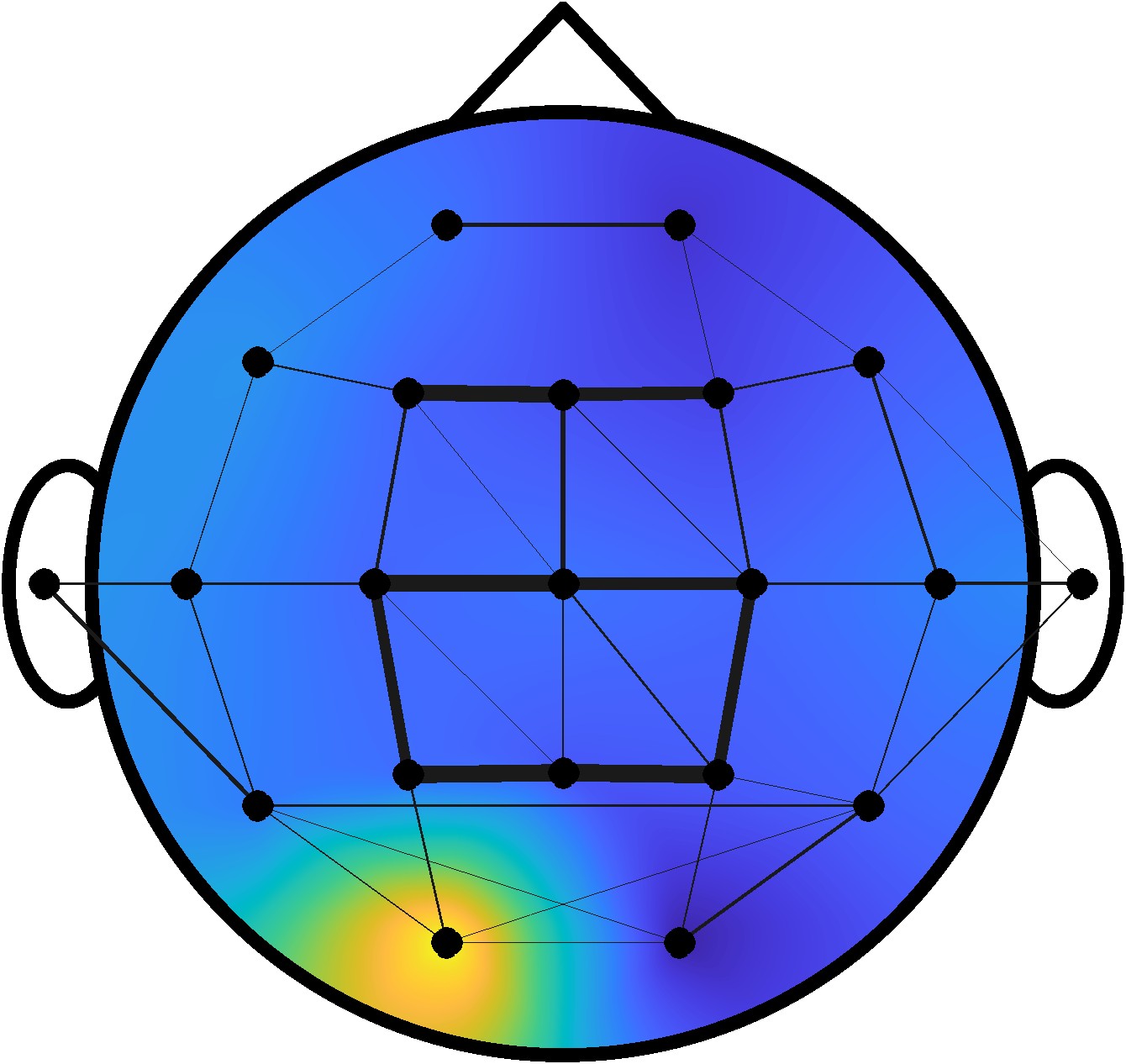}
        \caption{KSGL - normal}
    \end{subfigure}
    \hspace{0.01\textwidth}
    \begin{subfigure}[b]{0.2\textwidth}
        \centering
        \includegraphics[width=\textwidth]{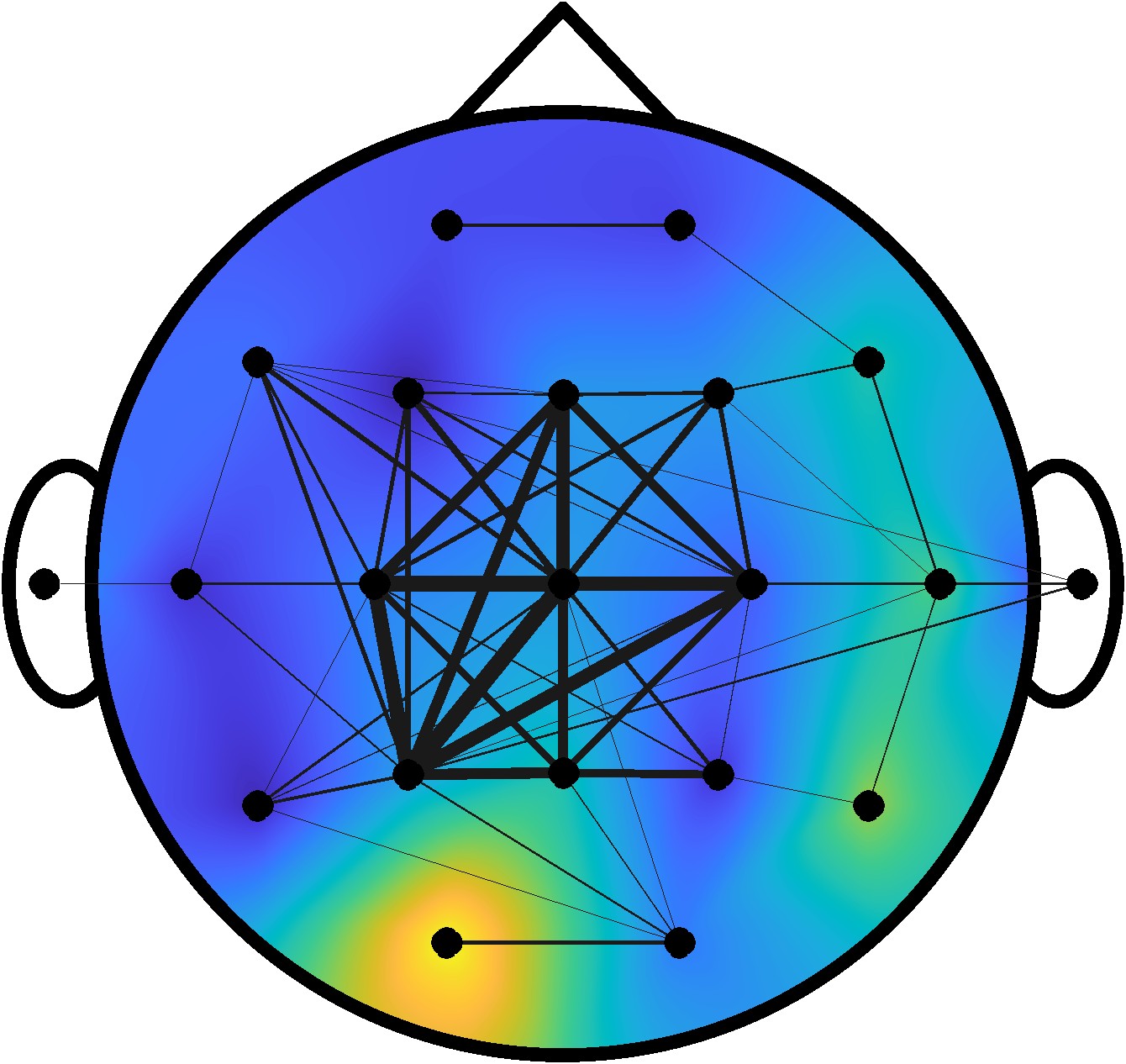}
        \caption{KSGL - seizure}
    \end{subfigure}
    \caption{The brain connectivity inferred by MWGL and KSGL. Nodes reflect the actual electrode positions in the 10-20 system. The background color shows the mean EEG activity of each status.
    }
    \label{fig:epileptic_brain_graphs}
\end{figure}

Because brain connectivity varies dramatically across individuals, we pick the EEG of a single patient to evaluate KSGL.
This 7-year-old male patient had several complex partial seizures in the central cortical area (Cz, C3, C4), but also went through multiple seizures that are not visible on EEG.
We apply KSGL to learn strong product graphs from his normal EEGs and deceptive epileptic EEGs then compare the results with those of MWGL.
Fig.~\ref{fig:epileptic_brain_graphs} shows the brain connectivity graphs learned by MWGL and KSGL.
As we can see, although the seizures are not obvious from the mean signal amplitude as expected, KSGL learns different connectivity patterns for these 2 statuses.
In particular, KSGL learns denser connectivity around the central cortical area, matching the known lesion.
On the other hand, MWGL learns similar brain connectivity patterns from normal and epileptic EEGs.
Also note that the brain connectivity graphs learned by KSGL are sparser than those by MWGL, suggesting that the strong product suits this dataset better than the Cartesian product.
Additional results of this patient and other patients are shown in Appendix~\ref{sec:addtional_eeg}.

\section{CONCLUSIONS}

In this paper, we focus on graph learning, a classical problem in GSP, and extend it to learning Kronecker product structures from multi-way signals.
We propose new algorithms for learning Kronecker and strong product graphs from smooth graph signals and evaluate their performance on both synthetic and real-world datasets.
Our experiments show that the proposed KSGL methods outperform competing GSP and GM methods.
We also investigate the theoretical aspects of the Kronecker algorithm and show that the solution of the penalized MLE converges to the true graph Laplacian asymptotically.
Our results also prove that KSGL converges faster than general graph learning where the product structures are ignored.
In the future, we intend to complete the theory for the strong product and improve the scalability of KSGL.



\newpage


\nocite{langley00}

\bibliography{icml2025}
\bibliographystyle{icml2025}

\newpage
\appendix
\onecolumn
\section{Proof of Main Theorems}

\subsection{Proof of Theorem~\ref{th:existence}}

\begin{proof}
    Given $\mathcal{A}\mathbf{w}=\mathcal{A}\mathbf{w}_1\otimes\mathcal{A}\mathbf{w}_2$, we now prove that the global minimizer of the penalized MLE
    \begin{equation}
    \label{eq:existence}
        \min_{\mathbf{w}_1,\mathbf{w}_2\geq \mathbf{0}}  \left\{ \langle \mathcal{L}\mathbf{w}, \mathbf{S} \rangle - \log{{\det}^\dagger(\mathcal{L}\mathbf{w})} + \alpha_1 {\|\mathbf{w}_1\|}_{1} + \alpha_2 {\|\mathbf{w}_2\|}_{1}  \right\},
    \end{equation}
    exists.
    Provided that both the product and factor graphs are connected, the feasible set over $\mathbf{w}_1$ and $\mathbf{w}_2$ is defined as
    \begin{equation}
        \Omega_{\mathbf{w}_1,\mathbf{w}_2} := \{(\mathbf{w}_1,\mathbf{w}_2)|\mathbf{w}_1\geq\mathbf{0}, \mathcal{L}\mathbf{w}_1+\mathbf{J}_{p_1}\in\mathbb{S}_{++}^{p_2}, \mathbf{w}_2\geq\mathbf{0}, \mathcal{L}\mathbf{w}_2+\mathbf{J}_{p_2}\in\mathbb{S}_{++}^{p_2}\},
    \end{equation}
    where $\mathbf{J}_p = \frac{1}{p}\mathbf{1}_p\mathbf{1}_p^T$ and we have ${\logdet}^\dagger(\mathcal{L}\mathbf{w})=\logdet(\mathcal{L}\mathbf{w}+\mathbf{J}_p)$.
    The conditions $\mathcal{L}\mathbf{w}_1+\mathbf{J}_{p_1}\in\mathbb{S}_{++}^{p_1}$ and $\mathcal{L}\mathbf{w}_2+\mathbf{J}_{p_2}\in\mathbb{S}_{++}^{p_2}$ constrain that $G_1$ and $G_2$ are connected.
    Let $\{0=\lambda_1<\lambda_2\leq\dots\leq\lambda_p\}$ be the eigenvalues of $\mathbf{L}=\mathcal{L}\mathbf{w}$. 
    We first consider the original MLE and bound the negative log-likelihood $Q(\mathbf{w}_1,\mathbf{w}_2)$ when $\alpha_1=\alpha_2=0$
    {\allowdisplaybreaks
    \begin{align}
        & \langle \mathcal{L}\mathbf{w}, \mathbf{S} \rangle-{\logdet}^\dagger(\mathcal{L}\mathbf{w})\label{eq:original_mle} \\
        = & \langle \mathcal{L}\mathbf{w}, \mathbf{S} \rangle-\log(\prod_{k=2}^{p}\lambda_k)\\
        \geq & \langle \mathcal{L}\mathbf{w}, \mathbf{S} \rangle-(p-1)\log(\sum_{k=1}^{p}\lambda_k) + (p-1)\log(p-1) \label{ieq:amgm}\\
        = & \langle \mathcal{L}^*\mathbf{S}, \mathbf{w} \rangle -(p-1)\log(p{\|\mathbf{w}\|}_1) + (p-1)\log(\frac{p-1}{2}) \label{ieq:trace}\\
        \geq & \min (\mathcal{L}^*\Tilde{\mathbf{S}}) p{\|\mathbf{w}\|}_1 -(p-1)\log(p{\|\mathbf{w}\|}_1) + (p-1)\log(\frac{p-1}{2}),
    \end{align}
    }
    where ${[\Tilde{\mathbf{S}}]}_{i,j}=\frac{1}{p}{[\mathbf{S}]}_{i,j}$.
    \eqref{ieq:trace} is attributed to the fact that the summation of eigenvalues equals the trace of the Laplacian.
    Define the function
    \begin{equation}
    \label{eq:mle_bd}
        q(t) = \min (\mathcal{L}^*\Tilde{\mathbf{S}}) t - (p-1)\log(t) + (p-1)\log(\frac{p-1}{2}).
    \end{equation}
    This function is lower-bounded at $t=\frac{p-1}{\min (\mathcal{L}^*\Tilde{\mathbf{S}})}$, so long as $\min (\mathcal{L}^*\Tilde{\mathbf{S}} )>0$.
    Therefore, we have that the negative log-likelihood is also lower-bounded
    \begin{equation}
        Q(\mathbf{w}_1,\mathbf{w}_2) \geq q(p{\|\mathbf{w}\|}_1) \geq (p-1)(1+\log(\frac{\min (\mathcal{L}^*\Tilde{\mathbf{S}})}{2})).
    \end{equation}
    We then notice that $q(t) \rightarrow \infty$ when $t \rightarrow \infty$.
    This is followed by $Q(\mathbf{w}_1,\mathbf{w}_2)$ being coercive.
    
    Now consider the penalized MLE.
    When the penalization $\alpha_1>0$ and $\alpha_2>0$, the penalized MLE $Q(\mathbf{w}_1,\mathbf{w}_2)$ is still lower-bounded. 
    Since 
    \begin{equation}
        \mathbf{w}_1\rightarrow \infty\ |\ \mathbf{w}_2\rightarrow \infty\ \leadsto\ Q(\mathbf{w}_1,\mathbf{w}_2)\rightarrow \infty,
    \end{equation}
    the penalized MLE $Q(\mathbf{w}_1,\mathbf{w}_2)$ is also coercive.
    Note that this only holds when we penalize the factor graphs.
    The penalized MLE wouldn't be coercive if the $\ell$-1 penalization is on the product graph.
    
    The above argument indicates that a global minimizer exists in $\mathrm{cl}(\Omega_{\mathbf{w}_1,\mathbf{w}_2})$, and now we show that it exist in $\Omega_{\mathbf{w}_1,\mathbf{w}_2}$.
    Since the open boundaries $\mathrm{cl}(\Omega_{\mathbf{w}_1,\mathbf{w}_2})\setminus \Omega_{\mathbf{w}_1,\mathbf{w}_2}$ are results of the connectivity constraint $\mathcal{L}\mathbf{w}_1+\mathbf{J}_{p_1}\succ \mathbf{O}$ and $\mathcal{L}\mathbf{w}_2+\mathbf{J}_{p_2}\succ \mathbf{O}$, we have that $\mathrm{cl}(\Omega_{\mathbf{w}_1,\mathbf{w}_2})\setminus \Omega_{\mathbf{w}_1,\mathbf{w}_2}$ is a subset of disconnected $\mathbf{w}_1$ and $\mathbf{w}_2$.
    The set of disconnected $\mathbf{w}_1$ and $\mathbf{w}_2$ is written as
    \begin{equation}
        \{(\mathbf{w}_1,\mathbf{w}_2)| \det(\mathcal{L}\mathbf{w}_1+\mathbf{J}_{p_1})=0 \vee \det(\mathcal{L}\mathbf{w}_2+\mathbf{J}_{p_2})=0 \}.
    \end{equation}
    Since for the Kronecker product, any factor graph being disconnected leads to the product graph being disconnected.
    Therefore,
    \begin{equation}
        (\mathbf{w}_1,\mathbf{w}_2)\in \mathrm{cl}(\Omega_{\mathbf{w}_1,\mathbf{w}_2})\setminus \Omega_{\mathbf{w}_1,\mathbf{w}_2}\ \leadsto\ \logdet(\mathcal{L}\mathbf{w}+\mathbf{J}_p)=-\infty\ \leadsto\ Q(\mathbf{w}_1,\mathbf{w}_2)\rightarrow\infty.
    \end{equation}
    This shows that any global minimizer over $\mathrm{cl}(\Omega_{\mathbf{w}_1,\mathbf{w}_2})$ do not lie on those open boundaries, therefore \eqref{eq:existence} has at least a global minimizer in $\Omega_{\mathbf{w}_1,\mathbf{w}_2}$ so long as $\min (\mathcal{L}^*\Tilde{\mathbf{S}})>0$, which almost surely holds with probability 1.
\end{proof}

\subsection{Proof of Theorem~\ref{th:uniqueness}}
\begin{proof}
    $Q(\mathbf{w}_1,\mathbf{w}_2)$ is not jointly convex on $\mathbf{w}_1$ and $\mathbf{w}_2$, but it is bi-convex with respect to each separate variable.
    This means the MLE objective is convex with respect to $\mathbf{w}_1$ when $\mathbf{w}_2$ is fixed, and also convex with respect to $\mathbf{w}_2$ when $\mathbf{w}_1$ is fixed.
    Define the feasible set of $\mathbf{w}_1$ and $\mathbf{w}_2$ as
    \begin{align}
        & \Omega_{\mathbf{w}_1}:=\{\mathbf{w}_1|\mathbf{w_1}>\mathbf{0},\mathcal{L}\mathbf{w}_1+\mathbf{J}_{p_1}\in\mathbb{S}_{++}^{p_1}\}\\
        & \Omega_{\mathbf{w}_2}:=\{\mathbf{w}_2|\mathbf{w_2}>\mathbf{0},\mathcal{L}\mathbf{w}_2+\mathbf{J}_{p_2}\in\mathbb{S}_{++}^{p_2}\}.
    \end{align}
    and we have $\Omega_{\mathbf{w}_1,\mathbf{w}_2}=\Omega_{\mathbf{w}_1}\times\Omega_{\mathbf{w}_2}$.
    To see that both $\Omega_{\mathbf{w}_1}$ and $\Omega_{\mathbf{w}_2}$ are convex sets,
    we check that $\forall\ \mathbf{w}_1^{0},\mathbf{w}_1^{1}\in\Omega_{\mathbf{w}_1}$ and $\forall\ \mathbf{w}_2^{0},\mathbf{w}_2^{1}\in\Omega_{\mathbf{w}_2}$
    \begin{align}
        & \mathcal{L}\mathbf{w}_1^{a}+\mathbf{J}_{p_1} = a (\mathcal{L}\mathbf{w}_1^{0}+\mathbf{J}_{p_1}) + (1-a) (\mathcal{L}\mathbf{w}_1^{1}+\mathbf{J}_{p_1}) \in \mathbb{S}_{++}^{p_1},\ \forall\ 0<a<1\\
        & \mathcal{L}\mathbf{w}_2^{b}+\mathbf{J}_{p_2} = b (\mathcal{L}\mathbf{w}_2^{0}+\mathbf{J}_{p_2}) + (1-b) (\mathcal{L}\mathbf{w}_2^{1}+\mathbf{J}_{p_2}) \in \mathbb{S}_{++}^{p_2},\ \forall\ 0<b<1,
    \end{align}
    where $\mathbf{w}_1^{a}=a\mathbf{w}_1^{0}+(1-a)\mathbf{w}_1^{1}>\mathbf{0}$ and $\mathbf{w}_2^{b}=b\mathbf{w}_2^{0}+(1-b)\mathbf{w}_2^{1}>\mathbf{0}$. 
    The set of positive definite matrices forms a convex cone.
    Now to prove $Q(\mathbf{w}_1,\mathbf{w}_2)$ is bi-convex, we again first consider the case where $\alpha_1=\alpha_2=0$.
    The negative log-likelihood \eqref{eq:original_mle} is convex with respect to $\mathbf{w}$.
    Because either $\mathbf{w}_1$ or $\mathbf{w}_2$ maps linearly to $\mathbf{w}$ when the other factor is fixed, \eqref{eq:original_mle} is bi-convex.
    The penalized MLE is then also bi-convex because both $\alpha_1{\|\mathbf{w}_1\|}_1$ and $\alpha_2{\|\mathbf{w}_2\|}_1$ are convex.
    Thus each sub-problem of the penalized MLE has a unique solution.
\end{proof}

\subsection{Proof of Corollary~\ref{cl:consistency}}

\begin{proof}
    We prove the first half of the corollary, then obtain the other half by symmetry.
    Let $\mathbf{L}^*$ be the Laplacian of the true Kronecker product graph to be estimated and $\mathcal{L}\mathbf{w}^*=\mathbf{L}^*$.
    By the properties of Kronecker graph product, $\mathbf{L}^*=\mathbf{D}^*-\mathbf{W}^*=\mathbf{D}_1^*\otimes\mathbf{D}_2^*-\mathbf{W}_1^*\otimes\mathbf{W}_2^*$.
    Let $\mathbf{L}_1^*$ and $\mathbf{L}_2^*$ be the true factor Laplacian, where $\mathcal{L}\mathbf{w}_1^*=\mathbf{L}_1^*$ and $\mathcal{L}\mathbf{w}_2^*=\mathbf{L}_2^*$.
    Although the factor Laplacians do not appear in the original problem formulation, they come in handy for deriving the consistency results.

    Let's define a set of perturbations around $\mathbf{L}_1^*$
    \begin{equation}
    \label{eq:delta_set}
        \mathcal{T}_1 = \{ \Delta_{\mathbf{L}_1}| \Delta_{\mathbf{L}_1} \in \mathcal{K}_{\mathbf{L}^*}, {\| \Delta_{\mathbf{L}_1} \|}_F = c_1r_{n,\mathbf{p}} \},
    \end{equation}
    where $r_{n,\mathbf{p}} = \sqrt{\frac{s_1p_1\log p}{np_2}}$ for $\mathbf{p}=(p,p_1,p_2)$ and 
    \begin{equation}
        \mathcal{K}_{\mathbf{L}_1^*} := \{\Delta_{\mathbf{L}_1}|\mathbf{L}_1^*+\Delta_{\mathbf{L}_1} \in \Omega_{\mathbf{L}_1} \}.
    \end{equation}
    If we can show that
    \begin{equation}
        F(\Delta_{\mathbf{L}_1}) = Q(\mathbf{L}_1^*+\Delta_{\mathbf{L}_1},\mathbf{L}_2^{\mathrm{init}}) - Q(\mathbf{L}_1^*,\mathbf{L}_2^{\mathrm{init}}) > 0,\ \forall \Delta_{\mathbf{L}_1} \in \mathcal{T}_1,
    \end{equation}
    then we will have
    \begin{equation}
    \label{ieq:ult_consistent}
        {\|\widehat{\mathbf{L}}_1-\mathbf{L}_1^*\|}_F \leq cr_{n,\mathbf{p}},
    \end{equation}
    following that $Q(\mathbf{L}_1,\mathbf{L}_2)$ is bi-convex, $F(\mathbf{O}_1)=0$, and $F(\widehat{\mathbf{L}}_1-\mathbf{L}_1^*) = Q(\widehat{\mathbf{L}}_1,\mathbf{L}_2^{\mathrm{init}})-Q(\mathbf{L}_1^*,\mathbf{L}_2^{\mathrm{init}}) \leq 0$.

    Let $\Delta_{\mathbf{L}}=\Delta_{\mathbf{D}_1}\otimes\mathbf{D}_2^{\mathrm{init}}-\Delta_{\mathbf{W}_1}\otimes\mathbf{W}_2^{\mathrm{init}}$.
    Then we can write
    \begin{equation}
        F(\Delta_{\mathbf{L}_1}) = \langle \Delta_\mathbf{L},\mathbf{S}\rangle - \left(\logdet(\mathbf{L}^*+\Delta_\mathbf{L}+\mathbf{J}_p) - \logdet(\mathbf{L}^*+\mathbf{J}_p ) \right) + \alpha_1 ({\|\mathbf{w}_1^*+\Delta_{\mathbf{w}_1}\|}_{1}-{\|\mathbf{w}_1^*\|}_{1}).
    \end{equation}
    Using Taylor's expansion of $\logdet(\mathbf{L}^*+\nu\Delta_\mathbf{L}+\mathbf{J}_p)$ with the integral remainder
    \begin{equation}
        \logdet(\mathbf{L}^*+\Delta_\mathbf{L}+\mathbf{J}_p) - \logdet(\mathbf{L}^*+\mathbf{J}_p) = \Tr({(\mathbf{L}^*+\mathbf{J}_p)}^{-1}\Delta_\mathbf{L}) + \int_0^1 (1-\nu) \nabla_\nu^2 \logdet(\mathbf{L}^*+\nu\Delta_\mathbf{L}+\mathbf{J}_p) d\nu,
    \end{equation}
    and further the remainder 
    \begin{align}
        \int_0^1 (1-\nu) \nabla_\nu^2 & \logdet(\mathbf{L}^*+\nu\Delta_\mathbf{L}+\mathbf{J}_p) d\nu = \\ \nonumber
       &  -{\mathrm{vec}(\Delta_\mathbf{L})}^T \left(\int_0^1 (1-\nu) {(\mathbf{L}^*+\nu\Delta_\mathbf{L}+\mathbf{J}_p)}^{-1} \otimes {(\mathbf{L}^*+\nu\Delta_\mathbf{L}+\mathbf{J}_p)}^{-1} d\nu \right) \mathrm{vec}(\Delta_\mathbf{L}).
    \end{align}
    Therefore we have
    \begin{align}
        F(\Delta_{\mathbf{L}_1}) = I_1 + I_2 + I_3,
    \end{align}
    where
    \begin{align}
        I_1 & = \langle \Delta_\mathbf{L}, \mathbf{S}-{(\mathbf{L}^*+\mathbf{J}_p)}^{-1} \rangle,\\
        I_2 & = {\mathrm{vec}(\Delta_\mathbf{L})}^T \left(\int_0^1 (1-\nu) {(\mathbf{L}^*+\nu\Delta_\mathbf{L}+\mathbf{J}_p)}^{-1} \otimes {(\mathbf{L}^*+\nu\Delta_\mathbf{L}+\mathbf{J}_p)}^{-1} d\nu\right) \mathrm{vec}(\Delta_\mathbf{L}),\\
        I_3 & = \alpha_1 ({\|\mathbf{w}_1^*+\Delta_{\mathbf{w}_1}\|}_{1}-{\|\mathbf{w}_1^*\|}_{1}).
    \end{align}

    \paragraph{Bound $I_1$:}

    The observations $\Bar{\mathbf{x}}$ are the samples from the improper GMRF \citep{ying2020nonconvex}
    \begin{equation}
        \Bar{\mathbf{x}} = \mathbf{x} -\frac{1}{p}\mathbf{1}\mathbf{1}^T\mathbf{x}, \;\; \mathbf{x}\sim\mathcal{N}(\mathbf{0},{(\mathbf{L}^*+\mathbf{J}_p)}^{-1}).
    \end{equation}
    Let $\mathbf{\Sigma}={(\mathbf{L}^*+\mathbf{J}_p)}^{-1}$ be the covariance matrix of the original proper GMRF. 
    Let $m_1=i-j+\frac{1}{2}(j-1)(2p_1-j),\ \forall 1\leq j<i\leq p$ and $\mathcal{L}\Delta_{\mathbf{w}} = \Delta_{\mathbf{L}}$, we have
    \begin{align}
        I_1 & = \Delta_{\mathbf{w}}^T\mathcal{L}^*(\mathbf{S}-{(\mathbf{L}^*+\mathbf{J}_p)}^{-1})\\
        & = \frac{1}{2} \langle \Delta_{\mathbf{W}_1} \otimes \mathbf{W}_2^{\mathrm{init}}, \mathcal{A}\mathcal{L}^*(\mathbf{S}-{(\mathbf{L}^*+\mathbf{J}_p)}^{-1}) \rangle\\
        & = \sum_{1\leq j<i\leq p_1}{[\Delta_{\mathbf{w}_1}]}_{m_1} \langle \mathcal{A}\mathbf{e}_{m_1} \otimes \mathbf{W}_2^{\mathrm{init}}, \mathcal{A}\mathcal{L}^*(\mathbf{S}-\mathbf{\Sigma}) \rangle \label{eq:quad_ref}\\
        & = \sum_{1\leq j<i\leq p_1}{[\Delta_{\mathbf{w}_1}]}_{m_1} \langle \mathbf{W}_2^{\mathrm{init}}, {[\mathbf{K}]}_{I_1,J_1} - {[\mathcal{A}\mathcal{L}^*\mathbf{\Sigma}]}_{I_1,J_1} \rangle\\
        & = \sum_{1\leq j<i\leq p_1}{[\Delta_{\mathbf{w}_1}]}_{m_1} \left( \langle \mathbf{W}_2^{\mathrm{init}}, {[\mathbf{K}]}_{I_1,J_1} \rangle - \langle \mathbf{W}_2^{\mathrm{init}}, {[\mathcal{A}\mathcal{L}^*\mathbf{\Sigma}]}_{I_1,J_1} \rangle \right),
    \end{align}
    where ${\mathbf{e}_{m_1}}\in\mathbb{R}^{\frac{p_1(p_1-1)}{2}}$ has $1$ in the $m_1$-th entry and $0$s otherwise.
    Also, notice that
    \begin{align}
        \mathbb{E}[{[\mathcal{L}^*\mathbf{S}]}_{m}] & = \mathbb{E}\left[\frac{1}{n}\sum_{k=1}^n{({[\Bar{\mathbf{x}_k}]}_i-{[\Bar{\mathbf{x}_k}]}_j)}^2\right]\\
        & = \frac{1}{n}\sum_{k=1}^n\mathbb{E}\left[{({[\mathbf{x}_k]}_i-{[\mathbf{x}_k]}_j)}^2\right]\\
        & = \frac{1}{n}\sum_{k=1}^n{\mathbb{E}[{[\mathbf{x}_k]}_i}^2] - 2\mathbb{E}[{[\mathbf{x}_k]}_i{[\mathbf{x}_k]}_j] + \mathbb{E}[{{[\mathbf{x}_k]}_j}^2]\\
        & = {[\mathbf{\Sigma}]}_{i,i}-{[\mathbf{\Sigma}]}_{i,j}-{[\mathbf{\Sigma}]}_{j,i}+{[\mathbf{\Sigma}]}_{j,j}\\
        & = {[\mathcal{L}^*\mathbf{\Sigma}]}_{m},
    \end{align}
    which leads to
    \begin{align}
        \mathbb{E}[\langle \mathbf{W}_2^{\mathrm{init}}, {[\mathbf{K}]}_{I_1,J_1} \rangle] = \langle \mathbf{W}_2^{\mathrm{init}}, \mathbb{E}[{[\mathbf{K}]}_{I_1,J_1}] \rangle 
        = \langle \mathbf{W}_2^{\mathrm{init}}, {[\mathcal{A}\mathcal{L}^*\mathbf{\Sigma}]}_{I_1,J_1} \rangle.
    \end{align}
    Therefore,
    \begin{align}
        I_1 = \sum_{1\leq j<i\leq p_1}{[\Delta_{\mathbf{w}_1}]}_{m_1} \left( \langle \mathbf{W}_2^{\mathrm{init}}, {[\mathbf{K}]}_{I_1,J_1} \rangle - \mathbb{E}[\langle \mathbf{W}_2^{\mathrm{init}}, {[\mathbf{K}]}_{I_1,J_1} \rangle] \right).
    \end{align}
    Now we bound the perturbation term.
    Let $\mathbf{x}_k=\mathbf{\Sigma}^{\frac{1}{2}}\mathbf{z}_k$, where $\mathbf{z}_k\sim\mathcal{N}(\mathbf{0},\mathbf{I}_p)$ is the source signal of the GSP system.
    From \eqref{eq:quad_ref} we know
    \begin{align}
        \langle \mathbf{W}_2^{\mathrm{init}}, {[\mathbf{K}]}_{I_1,J_1} \rangle & = \frac{1}{n} \sum_{k=1}^n \mathbf{x}_k^T (\mathcal{L}\mathcal{A}^*(\mathcal{A}\mathbf{e}_{m_1}\otimes\mathbf{W}_{2}^{\mathrm{init}})) \mathbf{x}_k \\
        & = \frac{1}{n} \sum_{k=1}^n {\mathbf{z}}_k^T \mathbf{\Sigma}^{\frac{1}{2}} (\mathcal{L}\mathcal{A}^*(\mathcal{A}\mathbf{e}_{m_1}\otimes\mathbf{W}_{2}^{\mathrm{init}})) \mathbf{\Sigma}^{\frac{1}{2}} \mathbf{z}_k. \label{eq:ls_quad}
    \end{align}
    Let $\mathbf{M}_{i,j} = \mathbf{\Sigma}^{\frac{1}{2}} (\mathcal{L}\mathcal{A}^*(\mathcal{A}\mathbf{e}_{m_1}\otimes\mathbf{W}_{2}^{\mathrm{init}})) \mathbf{\Sigma}^{\frac{1}{2}}$.
    We then apply the Hanson-Wright inequality \citep{hanson1971bound,rudelson2013hanson} to the quadratic
    \begin{align}
        & \mathbb{P} \left\{ \left|\frac{1}{n}\sum_{k=1}^n \mathbf{z}_k^T \mathbf{M}_{i,j} \mathbf{z}_k-\mathbb{E}\left[\frac{1}{n}\sum_{k=1}^n \mathbf{z}_k^T \mathbf{M}_{i,j} \mathbf{z}_k\right]\right|>h \right\}\\
        \leq & 2\exp{\left[-c_1' \min{\left(\frac{nh^2}{K^4{\|\mathbf{M}_{i,j}\|}_F^2},\frac{nh}{K^2{\|\mathbf{M}_{i,j}\|}_2}\right)}\right]} \label{ieq:hanson_wright}\\
        \leq & 2\exp{\left[-c_1' \min{\left(\frac{nh^2}{K^4{\|\mathbf{M}_{i,j}\|}_F^2},\frac{nh}{K^2{\|\mathbf{M}_{i,j}\|}_F}\right)}\right]} \label{ieq:fro_property} \\
        \leq & 2\exp{\left[-c_1' \min{\left(\frac{nh^2}{2K^4{\|\mathcal{L}\mathbf{w}_2^{\mathrm{init}}\|}_F^2{\|\mathbf{\Sigma}\|}_2^2},\frac{nh}{\sqrt{2}K^2{\|\mathcal{L}\mathbf{w}_{2}^{\mathrm{init}}\|}_F{\|\mathbf{\Sigma}\|}_2}\right)}\right]} \label{ieq:spectral_norm}\\
        \leq & 2\exp{\left[-c_1' \min{\left(\frac{nh^2}{32{\|\mathcal{L}\mathbf{w}_2^{\mathrm{init}}\|}_F^2{\|\mathbf{\Sigma}\|}_2^2},\frac{nh}{4\sqrt{2}{\|\mathcal{L}\mathbf{w}_{2}^{\mathrm{init}}\|}_F{\|\mathbf{\Sigma}\|}_2}\right)}\right]}\label{ieq:sub_gaussian}, 
    \end{align}
    where from \eqref{ieq:fro_property} to \eqref{ieq:spectral_norm} we 
    use
    \begin{align}
        & {\|\mathbf{M}_{i,j}\|}_F^2 \leq {\|\mathbf{\Sigma}^{\frac{1}{2}}\|}_2^2 {\|\mathcal{L}\mathcal{A}^*(\mathcal{A}\mathbf{e}_{m_1}\otimes\mathbf{W}_{2}^{\mathrm{init}})\|}_F^2 {\|\mathbf{\Sigma}^{\frac{1}{2}}\|}_2^2 = 2{\|\mathcal{L}\mathbf{w}_{2}^{\mathrm{init}}\|}_F^2{\|\mathbf{\Sigma}\|}_2^2,
    \end{align}
    and from \eqref{ieq:spectral_norm} to \eqref{ieq:sub_gaussian} we use $K\leq2$ for the sub-Gaussian norm of $\mathbf{z}_k$.
    Let $\epsilon=\frac{h}{{\|\mathcal{L}\mathbf{w}_{2}^{\mathrm{init}}\|}_F{{\|\mathbf{\Sigma}\|}_2}}$ and plug \eqref{eq:ls_quad} into \eqref{ieq:sub_gaussian}, we arrive at
    \begin{equation}
        \mathbb{P} \left\{ |\langle \mathbf{W}_2^{\mathrm{init}}, {[\mathbf{K}]}_{I_1,J_1} \rangle-\mathbb{E}[\langle \mathbf{W}_2^{\mathrm{init}}, {[\mathbf{K}]}_{I_1,J_1} \rangle]|>\epsilon {\|\mathcal{L}\mathbf{w}_{2}^{\mathrm{init}}\|}_F{\|\mathbf{\Sigma}\|}_2 \right\}  \leq 2\exp{\left(-\frac{c_1'n\epsilon^2}{32}\right)}, \ \forall \epsilon\leq4\sqrt{2}.
    \end{equation}
    The union bound indicates that
    \begin{align}
        & \mathbb{P} \left\{\max_{m_1}{\left[ |\langle \mathbf{W}_2^{\mathrm{init}}, {[\mathbf{K}]}_{I_1,J_1} \rangle-\mathbb{E}[\langle \mathbf{W}_2^{\mathrm{init}}, {[\mathbf{K}]}_{I_1,J_1} \rangle]| \right]} > \epsilon {\|\mathcal{L}\mathbf{w}_{2}^{\mathrm{init}}\|}_F{\|\mathbf{\Sigma}\|}_2 \right\} \\
        \leq & \sum_{m_1=1}^{\frac{p_1(p_1-1)}{2}} 2\exp{\left(-\frac{c_1'n\epsilon^2}{32}\right)} \\
        \leq & p_1^2\exp{\left(-\frac{c_1'n\epsilon^2}{32}\right)},
    \end{align}
    so
    \begin{align}
        \mathbb{P} \left\{\max_{m_1}{\left[ |\langle \mathbf{W}_2^{\mathrm{init}}, {[\mathbf{K}]}_{I_1,J_1} \rangle-\mathbb{E}[\langle \mathbf{W}_2^{\mathrm{init}}, {[\mathbf{K}]}_{I_1,J_1} \rangle]| \right]} \leq \epsilon {\|\mathcal{L}\mathbf{w}_{2}^{\mathrm{init}}\|}_F{\|\mathbf{\Sigma}\|}_2 \right\} \geq 1-p_1^2\exp{\left(-\frac{c_1'n\epsilon^2}{32}\right)}.
    \end{align}
    Thus with the above probability and $\epsilon\leq 4\sqrt{2}$
    \begin{align}
        I_1 & = \sum_{1\leq j<i\leq p_1}{[\Delta_{\mathbf{w}_1}]}_{m_1} \left( \langle \mathbf{W}_2^{\mathrm{init}}, {[\mathbf{K}]}_{I_1,J_1} \rangle - \mathbb{E}[\langle \mathbf{W}_2^{\mathrm{init}}, {[\mathbf{K}]}_{I_1,J_1} \rangle] \right)\\
        & \geq -\max_{m_1}{\left[ |\langle \mathbf{W}_2^{\mathrm{init}}, {[\mathbf{K}]}_{I_1,J_1} - {[\mathcal{A}\mathcal{L}^*\mathbf{\Sigma}]}_{I_1,J_1} \rangle| \right]}{\|\Delta_{\mathbf{w}_1}\|}_1\\
        & \geq -\epsilon {\|\mathcal{L}\mathbf{w}_{2}^{\mathrm{init}}\|}_F{\|\mathbf{\Sigma}\|}_2 {\|\Delta_{\mathbf{w}_1}\|}_1\\
        & \geq -\epsilon \sqrt{p_2}{\|\mathcal{L}\mathbf{w}_{2}^{\mathrm{init}}\|}_2{\|\mathbf{\Sigma}\|}_2 {\|\Delta_{\mathbf{w}_1}\|}_1.
    \end{align}

    \paragraph{Bound $I_2$:}

    From the min-max theorem, we have
    \begin{equation}
        I_2 \geq {\|\Delta_\mathbf{L}\|}_F^2 \lambda_{\mathrm{min}} \left(\int_0^1 (1-\nu) {(\mathbf{L}^*+\nu\Delta_\mathbf{L}+\mathbf{J}_p)}^{-1} \otimes {(\mathbf{L}^*+\nu\Delta_\mathbf{L}+\mathbf{J}_p)}^{-1} d\nu\right).
    \end{equation}
    Then given the convexity of $\lambda_{\mathrm{max}}(\cdot)$ and concavity of $\lambda_{\mathrm{min}}(\cdot)$
    {\allowdisplaybreaks
    \begin{align}
        & \lambda_{\mathrm{min}} \left(\int_0^1 (1-\nu) {(\mathbf{L}^*+\nu\Delta_\mathbf{L}+\mathbf{J}_p)}^{-1} \otimes {(\mathbf{L}^*+\nu\Delta_\mathbf{L}+\mathbf{J}_p)}^{-1} d\nu\right) \\
        \geq & \int_0^1 (1-\nu) \lambda_{\mathrm{min}}^2{(\mathbf{L}^*+\nu\Delta_\mathbf{L}+\mathbf{J}_p)}^{-1} d\nu \\
        \geq & \min_{\nu\in [0,1]} \left[ \lambda_{\mathrm{min}}^2{(\mathbf{L}^*+\nu\Delta_\mathbf{L}+\mathbf{J}_p)}^{-1}\right] \int_0^1 (1-\nu) d\nu \\
        = & \frac{1}{2} \min_{\nu\in [0,1]} \left[ \frac{1}{\|\mathbf{L}^*+\nu\Delta_\mathbf{L}+\mathbf{J}_p\|}_2^2 \right] \\
        = & \frac{1}{2\max_{\nu\in [0,1]} \left[ {\|\mathbf{L}^*+\nu\Delta_\mathbf{L}+\mathbf{J}_p\|}_2^2\right] } \\
        \geq & \frac{1}{2\max_{\nu\in [0,1]}^2 {\left[{\|\mathbf{L}^*+\mathbf{J}_p\|}_2+{\|\nu\Delta_\mathbf{L}\|}_2\right]}} \\
        = & \frac{1}{2{({\|\mathbf{L}^*+\mathbf{J}_p\|}_2+{\|\Delta_\mathbf{L}\|}_2)}^2}
    \end{align}
    }
    Let $\mathbf{d}_1$ denotes the diagonal of $\mathbf{D}_1$. The Gershgorin circle theorem implies that
    \begin{equation}
        {\|\Delta_\mathbf{L}\|}_2 \leq 2{\|\Delta_{\mathbf{d}_1}\|}_{\infty}d_{2,\mathrm{max}} \leq 2d_{2,\mathrm{max}}{\|\Delta_{\mathbf{L}_1}\|}_F = 2cd_{2,\mathrm{max}}r_{n,\mathbf{p}}
    \end{equation}
    Then with $n$ sufficiently large $n \geq \frac{4c^2 d_{2,\mathrm{max}}^2s_1p_1\log p}{p_2{\|\mathbf{L}^*+\mathbf{J}_p\|}_2^2}$ such that ${\|\Delta_\mathbf{L}\|}_2 \leq 2cd_{2,\mathrm{max}}r_{n,\mathbf{p}} \leq {\|\mathbf{L}^*+\mathbf{J}_p\|}_2 $,
    we obtain
    \begin{align}
        I_2 & \geq \frac{{\|\Delta_\mathbf{L}\|}_F^2}{8 {\|\mathbf{L}^*+\mathbf{J}_p\|}_2^2}.
    \end{align}
    To factor $\Delta_{\mathbf{L}_1}$ out, note that
    \begin{align}
        {\|\Delta_\mathbf{L}\|}_F^2 
        & = {\|\Delta_{\mathbf{W}_1}\otimes\mathbf{W}_2^{\mathrm{init}}\|}_F^2 + {\|\Delta_{\mathbf{D}_1}\otimes\mathbf{D}_2^{\mathrm{init}}\|}_F^2\\
        & = {\|\Delta_{\mathbf{W}_1}\|}_F^2{\|\mathbf{W}_2^{\mathrm{init}}\|}_F^2 + {\|\Delta_{\mathbf{D}_1}\|}_F^2{\|\mathbf{D}_2^{\mathrm{init}}\|}_F^2\\
        & \geq {\|\Delta_{\mathbf{L}_1}\|}_F^2{\|\mathbf{W}_2^{\mathrm{init}}\|}_F^2, \label{ieq:bound_deltaL1}
    \end{align}
    and that $\|\mathbf{W}_2^{\mathrm{init}}\|$ is lower and upper bounded
    \begin{align}
        \sqrt{\frac{p_2}{s_2}}d_{2,\mathrm{min}} \leq {\|\mathbf{W}_2^{\mathrm{init}}\|}_F \leq \sqrt{p_2}d_{2,\mathrm{max}}
    \end{align}
    Thus
    \begin{equation}
        I_2 \geq \frac{p_2d_{2,\mathrm{min}}^2{\|\Delta_{\mathbf{L}_1}\|}_F^2}{8 s_2{\|\mathbf{L}^*+\mathbf{J}_p\|}_2^2}.
    \end{equation}

    \paragraph{Bound $I_3$:}
    With the triangle inequality
    \begin{align}
        {\|\mathbf{w}_1^*+\Delta_{\mathbf{w}_1}\|}_1-{\|\mathbf{w}_1^*\|}_1 = {\|\mathbf{w}_1^*+\Delta_{\mathbf{w}_1}\|}_{1,\mathcal{S}_1}+{\|\Delta_{\mathbf{w}_1}\|}_{1,\mathcal{S}_1^\complement}-{\|\mathbf{w}_1^*\|}_{1,\mathcal{S}_1}
        \geq {\|\Delta_{\mathbf{w}_1}\|}_{1,\mathcal{S}_1^\complement}-{\|\Delta_{\mathbf{w}_1}\|}_{1,\mathcal{S}_1},
    \end{align}
    we can lower-bound $I_3$
    \begin{equation}
        I_3 \geq 
        2\alpha_1{\|\Delta_{\mathbf{w}_1}\|}_{1,\mathcal{S}_1^\complement} - 2\alpha_1{\|\Delta_{\mathbf{w}_1}\|}_{1,\mathcal{S}_1}.
    \end{equation}

    \paragraph{Bound $I_1+I_2+I_3$:}

    Overall,
    \begin{align}
        F(\Delta_{\mathbf{L}_1}) & \geq -\epsilon \sqrt{p_2}{\|\mathcal{L}\mathbf{w}_{2}^{\mathrm{init}}\|}_2{\|\mathbf{\Sigma}\|}_2 {\|\Delta_{\mathbf{w}_1}\|}_1 + \frac{{\|\Delta_\mathbf{L}\|}_F^2}{8 {\|\mathbf{L}^*+\mathbf{J}_p\|}_2^2} + 2\alpha_1{\|\Delta_{\mathbf{w}_1}\|}_{1,\mathcal{A}_1^\complement} - 2\alpha_1{\|\Delta_{\mathbf{w}_1}\|}_{1,\mathcal{A}_1}\\
        & = \frac{{\|\Delta_\mathbf{L}\|}_F^2}{8 {\|\mathbf{L}^*+\mathbf{J}_p\|}_2^2} - (\epsilon \sqrt{p_2}{\|\mathcal{L}\mathbf{w}_{2}^{\mathrm{init}}\|}_2{\|\mathbf{\Sigma}\|}_2 -2\alpha_1){\|\Delta_{\mathbf{w}_1}\|}_{1,\mathcal{A}_1^\complement} - (\epsilon \sqrt{p_2}{\|\mathcal{L}\mathbf{w}_{2}^{\mathrm{init}}\|}_2{\|\mathbf{\Sigma}\|}_2 +2\alpha_1){\|\Delta_{\mathbf{w}_1}\|}_{1,\mathcal{A}_1}.
    \end{align}
    Let $\epsilon=c_1''\sqrt{\frac{\log p}{n}}$ with sufficiently large
    \begin{equation}
        n \geq \frac{c_1''^2\log p}{32},
    \end{equation}
    so that $\epsilon \leq 4\sqrt{2}$ is satisfied.
    Then choose
    \begin{equation}
        \alpha_1 \geq \frac{c_1''{\|\mathcal{L}\mathbf{w}_{2}^{\mathrm{init}}\|}_2{\|\mathbf{\Sigma}\|}_2}{2}\sqrt{\frac{p_2\log p}{n}}, \label{ieq:alpha1_range}
    \end{equation}
    so that $\epsilon \sqrt{p_2}{\|\mathcal{L}\mathbf{w}_{2}\|}_2{\|\mathbf{\Sigma}\|}_2 -2\alpha<0$ and
    \begin{align}
        F(\Delta_{\mathbf{L}_1}) \geq \frac{{\|\Delta_\mathbf{L}\|}_F^2}{8 {\|\mathbf{L}^*+\mathbf{J}_p\|}_2^2} - (\epsilon \sqrt{p_2}{\|\mathcal{L}\mathbf{w}_{2}^{\mathrm{init}}\|}_2{\|\mathbf{\Sigma}\|}_2 +2\alpha_1){\|\Delta_{\mathbf{w}_1}\|}_{1,\mathcal{A}_1},
    \end{align}
    We can also bound ${\|\Delta_{\mathbf{w}_1}\|}_{1,\mathcal{A}_1}$ by
    \begin{equation}
        {\|\Delta_{\mathbf{w}_1}\|}_{1,\mathcal{S}_1} \leq \sqrt{p_1s_1} {\|\Delta_{\mathbf{w}_1}\|}_{2,\mathcal{S}_1} \leq\sqrt{p_1s_1} {\|\Delta_{\mathbf{w}_1}\|}_2 \leq \sqrt{\frac{p_1s_1}{2}}{\|\Delta_{\mathbf{L}_1}\|}_F.\label{ieq:bound_p1}
    \end{equation}
    Now, to prove $F(\Delta_{\mathbf{L}_1}) \geq 0$, define a ratio factor $\gamma_1\geq1$ that controls the $\ell_1$ penalty $\frac{\alpha_1}{\gamma_1} = \frac{c_1''{\|\mathcal{L}\mathbf{w}_{2}^{\mathrm{init}}\|}_2{\|\mathbf{\Sigma}\|}_2}{2}\sqrt{\frac{p_2\log p}{n}}$. 
    We obtain
    \begin{align}
        F(\Delta_{\mathbf{L}_1}) & \geq \frac{{\|\Delta_\mathbf{L}\|}_F^2}{8 {\|\mathbf{L}^*+\mathbf{J}_p\|}_2^2} - (1+\gamma_1)c_1''{\|\mathcal{L}\mathbf{w}_{2}^{\mathrm{init}}\|}_2{\|\mathbf{\Sigma}\|}_2\sqrt{\frac{p_2\log p}{n}}{\|\Delta_{\mathbf{w}_1}\|}_{1,\mathcal{S}_1}\\
        & \geq p_2{\|\Delta_{\mathbf{L}_1}\|}_F^2\left(\frac{d_{2,\mathrm{min}}^2}{8s_2 {\|\mathbf{L}^*+\mathbf{J}_p\|}_2^2} - (1+\gamma_1)c_1''{\|\mathcal{L}\mathbf{w}_{2}^{\mathrm{init}}\|}_2{\|\mathbf{\Sigma}\|}_2\sqrt{\frac{s_1p_1\log p}{2p_2n}}{\|\Delta_{\mathbf{L}_1}\|}_F^{-1}\right)\\
        & = p_2{\|\Delta_{\mathbf{L}_1}\|}_F^2\left(\frac{d_{2,\mathrm{min}}^2}{8s_2 {\|\mathbf{L}^*+\mathbf{J}_p\|}_2^2} - (1+\gamma_1)\frac{c_1''{\|\mathcal{L}\mathbf{w}_2^{\mathrm{init}}\|}_2{\|\mathbf{\Sigma}\|}_2}{\sqrt{2}c_1}\right)\\
        & > 0,
    \end{align}
    for sufficiently large $c$
    \begin{equation}
        c_1 \geq 4\sqrt{2}(1+\gamma_1)\frac{c_1''s_2}{d_{2,\mathrm{min}}^2}{\|\mathcal{L}\mathbf{w}_2^{\mathrm{init}}\|}_2{\|\mathbf{\Sigma}\|}_2{\|\mathbf{L}^*+\mathbf{J}_p\|}_2^2.
    \end{equation}
    This happens with probability
    \begin{align}
        & \mathbb{P} \left\{\max_{m_1}{\left[ |\langle \mathbf{W}_2^{\mathrm{init}}, {[\mathbf{K}]}_{I_1,J_1} \rangle-\mathbb{E}[\langle \mathbf{W}_2^{\mathrm{init}}, {[\mathbf{K}]}_{I_1,J_1} \rangle]| \right]} \leq \epsilon {\|\mathcal{L}\mathbf{w}_{2}^{\mathrm{init}}\|}_F{\|\mathbf{\Sigma}\|}_2 \right\}\\
        \geq & 1-p_1^2\exp{(-\frac{c_1'n\epsilon^2}{32})} \\
        = & 1-\exp{\left[2\log p_1-\frac{c_1'c_1''^2}{32}\log p\right]}\\
        \geq &1-\exp{(-(\frac{c_1'c_1''^2}{32}-2)\log p)}.
    \end{align}

    We have proved that the $\mathbf{w}_1$ estimation is consistent when fixing $\mathbf{w}_2$.
\end{proof}

\subsection{Proof of Theorem~\ref{th:consistency}}
\begin{proof}
    The algorithm starts with fixing $\mathbf{w}_2=\mathbf{w}_2^{\mathrm{init}}$ and updating $\mathbf{w}_1$, whose error is obtained by Corollary~\ref{cl:consistency}.
    Now we move forward to prove the consistency of $\mathbf{w}_2$ when fixing $\widehat{\mathbf{w}}_1=\mathbf{w}_1^{(1)}$.
    Similarly, we aim to show
    \begin{equation}
        G(\Delta_{\mathbf{L}_2}) = Q(\mathbf{L}_1^{(1)},\mathbf{L}_2^*+\Delta_{\mathbf{L}_2}) - Q(\mathbf{L}_1^{(1)},\mathbf{L}_2^*) > 0,\ \forall \Delta_{\mathbf{L}_2} \in \mathcal{T}_2,
    \end{equation}
    \begin{equation}
        \mathcal{T}_2 = \{ \Delta_{\mathbf{L}_2}| \Delta_{\mathbf{L}_2} \in \mathcal{K}_{\mathbf{L}^*}, {\| \Delta_{\mathbf{L}_2} \|}_F = c_2r_{n,\mathbf{p}} \},
    \end{equation}
    where $r_{n,\mathbf{p}} = \sqrt{\frac{s_2p_2\log p}{np_1}}$.
    By symmetry, with high probability, for $\epsilon=c_2''\sqrt{\frac{\log p}{n}}$ and $n \geq \frac{c_2''^2\log p}{32}$
    \begin{align}
        G(\Delta_{\mathbf{L}_2}) & \geq -\epsilon \sqrt{p_1}{\|\mathcal{L}\mathbf{w}_{1}^{(1))}\|}_2{\|\mathbf{\Sigma}\|}_2 {\|\Delta_{\mathbf{w}_2}\|} + \frac{{\|\Delta_\mathbf{L}\|}_F^2}{8 {\|\mathbf{L}^*+\mathbf{J}_p\|}_2^2} + 2\alpha_2{\|\Delta_{\mathbf{w}_2}\|}_{1,\mathcal{A}_2^\complement} - 2\alpha_2{\|\Delta_{\mathbf{w}_2}\|}_{1,\mathcal{A}_2}\\
        & = \frac{{\|\Delta_\mathbf{L}\|}_F^2}{8 {\|\mathbf{L}^*+\mathbf{J}_p\|}_2^2} - (\epsilon \sqrt{p_1}{\|\mathcal{L}\mathbf{w}_{1}^{(1)}\|}_2{\|\mathbf{\Sigma}\|}_2 -2\alpha_2){\|\Delta_{\mathbf{w}_2}\|}_{1,\mathcal{A}_2^\complement} - (\epsilon \sqrt{p_1}{\|\mathcal{L}\mathbf{w}_{1}^{(1)}\|}_2{\|\mathbf{\Sigma}\|}_2 +2\alpha_2){\|\Delta_{\mathbf{w}_2}\|}_{1,\mathcal{A}_2},
    \end{align}
    where $\Delta_{\mathbf{L}}=\mathbf{D}_1^{(1)}\otimes\Delta_{\mathbf{D}_2}-\mathbf{W}_1^{(1)}\otimes\Delta_{\mathbf{W}_2}$.
    Different from the previous derivation, $\mathbf{L}_1^{(1)}$ and $\mathbf{W}_1^{(1)}$ are now variables and we use ${\|\mathbf{L}_1^{(1)}-\mathbf{L}_1^*\|}_F \leq c\sqrt{\frac{s_1p_1\log p}{np_2}}$ to bound them.
    First ,we have
    \begin{align}
        & {\|\mathbf{W}_1^{(1)}\|}_F^2 \geq {\|\mathbf{W}_1^*\|}_F^2 - 2{\|\mathbf{W}_1^*\|}_F{\|\mathbf{W}_1^{(1)}-\mathbf{W}_1^*\|}_F \geq \frac{d_{1,\mathrm{min}}^2p_1}{s_1} - 2c_1d_{1,\mathrm{max}}p_1\sqrt{\frac{s_1\log p}{np_2}}.
    \end{align}
    Then, let $d_{1,\mathrm{max}}^{(1)}$ be the maximum degree of $\mathbf{W}_1^{(1)}$, by the Gershgorin circle theorem
    \begin{align}
        {\|\mathbf{L}_1^{(1)}\|}_2 \leq 2d_{1,\mathrm{max}}^{(1)} \leq 2d_{1,\mathrm{max}} + 2c_1\sqrt{\frac{s_1p_1\log p}{np_2}}.
    \end{align}
    Similar to \eqref{ieq:alpha1_range}, we choose a large enough $\alpha_2$    with a ratio factor $\gamma_2\geq1$
    \begin{equation}
        \frac{\alpha_2}{\gamma_2}=c_2''\left(d_{1,\mathrm{max}} + c_1\sqrt{\frac{s_1p_1\log p}{np_2}}\right){\|\mathbf{\Sigma}\|}_2\sqrt{\frac{p_1\log p}{n}}.\label{eq:alpha2_factor}
    \end{equation} 
    
    Similar to \eqref{ieq:bound_deltaL1} and \eqref{ieq:bound_p1}, we have
    \begin{align}
        {\|\Delta_\mathbf{L}\|}_F^2 \geq {\|\Delta_{\mathbf{L}_2}\|}_F^2{\|\mathbf{W}_1^{(1))}\|}_F^2,\label{ieq:bound_deltaL2}
    \end{align}
    and
    \begin{equation}
        {\|\Delta_{\mathbf{w}_2}\|}_{1,\mathcal{S}_2} \leq \sqrt{p_2s_2} {\|\Delta_{\mathbf{w}_2}\|}_{2,\mathcal{S}_2} \leq\sqrt{p_2s_2} {\|\Delta_{\mathbf{w}_2}\|}_2 \leq \sqrt{\frac{p_2s_2}{2}}{\|\Delta_{\mathbf{L}_2}\|}_F.\label{ieq:bound_p2}
    \end{equation}
    Plugging in $\epsilon=c_2''\sqrt{\frac{\log p}{n}}$, \eqref{eq:alpha2_factor}, \eqref{ieq:bound_deltaL2}, and \eqref{ieq:bound_p2}, we obtain

    \begin{align}
        G(\Delta_{\mathbf{L}_2}) \geq & {\|\Delta_{\mathbf{L}_2}\|}_F^2 \left(\frac{{\|\mathbf{W}_1^{(1)}\|}_F^2}{8 {\|\mathbf{L}^*+\mathbf{J}_p\|}_2^2} - (1+\gamma_2)c_2'' \sqrt{\frac{s_2p_2\log p}{2np_1}}{\|\mathcal{L}\mathbf{w}_{1}^{(1))}\|}_2{\|\mathbf{\Sigma}\|}_2{\|\Delta_{\mathbf{L}_2}\|}_F^{-1}\right)\\
        \geq & p_1{\|\Delta_{\mathbf{L}_2}\|}_F^2\left( \frac{1}{8 {\|\mathbf{L}^*+\mathbf{J}_p\|}_2^2}\left(\frac{d_{1,\mathrm{min}}^2}{s_1 } - 2c_1d_{1,\mathrm{max}}\sqrt{\frac{s_1\log p}{np_2}}\right) \right.\nonumber\\
        &\qquad\qquad\qquad\qquad\qquad\qquad \left. - \sqrt{2}(1+\gamma_2)\frac{c_2''}{c_2} \left(d_{1,\mathrm{max}} + c_1\sqrt{\frac{s_1p_1\log p}{np_2}}\right){\|\mathbf{\Sigma}\|}_2\right)\\
        \geq & p_1{\|\Delta_{\mathbf{L}_2}\|}_F^2\left( \frac{(1-\zeta)d_{1,\mathrm{min}}^2}{8 s_1{\|\mathbf{L}^*+\mathbf{J}_p\|}_2^2}  - \sqrt{2}(1+\gamma_2)(1+\iota)\frac{c_2''d_{1,\mathrm{max}}}{c_2} {\|\mathbf{\Sigma}\|}_2\right)\\
        \geq & 0,
    \end{align}
    for sufficiently large $n$
    \begin{equation}
        n \geq \max\left[ \frac{2c_1d_{1,\mathrm{max}}s_1^2\log p}{\zeta^2d_{1,\mathrm{min}}^4p_2}, \frac{c_1^2s_1p_1\log p}{\iota^2d_{1,\mathrm{max}}^2p_2}\right],
    \end{equation}
    and $c_2$
    \begin{equation}
        c_2\geq 8\sqrt{2}\frac{(1+\gamma)(1+\iota)c_2''d_{1,\mathrm{max}}s_1}{(1-\zeta)d_{1,\mathrm{min}}^2}{\|\mathbf{\Sigma}\|}_2{\|\mathbf{L}^*+\mathbf{J}_p\|}_2^2,
    \end{equation}
    where $0<\zeta<1$ and $0<\iota$ are additional ratio factors.
    We have now proved that the second iteration of the alternating optimization is consistent.
    With induction, one can show that
    \begin{align}
        & {\|\mathbf{L}_1^{(2t-1)}-\mathbf{L}_1^*\|}_F \leq c_1\sqrt{\frac{s_1p_1\log p}{np_2}},\\
        & {\|\mathbf{L}_2^{(2t)}-\mathbf{L}_2^*\|}_F \leq c_2\sqrt{\frac{s_2p_2\log p}{np_1}}.
    \end{align}

    Finally, to show the convergence of the product graphs, we decompose the error as
    \begin{align}
        \Delta_{\mathbf{W}^{(2t)}} & = \mathbf{W}_1^{(2t-1)}\otimes\mathbf{W}_2^{(2t)} - \mathbf{W}_1^*\otimes\mathbf{W}_2^* \\
        & = \mathbf{W}_1^*\otimes\Delta_{\mathbf{W}_2^{(2t)}} + \Delta_{\mathbf{W}_1^{(2t-1)}}\otimes\mathbf{W}_2^* + \Delta_{\mathbf{W}_1^{(2t-1)}}\otimes\Delta_{\mathbf{W}_2^{(2t)}},
    \end{align}
    from which we obtain
    \begin{align}
        {\|\Delta_{\mathbf{W}^{(2t)}}\|}_F & \leq {\|\mathbf{W}_1^*\otimes\Delta_{\mathbf{W}_2^{(2t)}}\|}_F + {\|\Delta_{\mathbf{W}_1^{(2t-1)}}\otimes\mathbf{W}_2^*\|}_F + {\|\Delta_{\mathbf{W}_1^{(2t-1)}}\otimes\Delta_{\mathbf{W}_2^{(2t)}}\|}_F\\
        & = {\|\mathbf{W}_1^*\|}_F{\|\Delta_{\mathbf{W}_2^{(2t)}}\|}_F + {\|\Delta_{\mathbf{W}_1^{(2t-1)}}\|}_F{\|\mathbf{W}_2^*\|}_F + {\|\Delta_{\mathbf{W}_1^{(2t-1)}}\|}_F{\|\Delta_{\mathbf{W}_2^{(2t)}}\|}_F\\
        & \leq c_1d_{1,\mathrm{max}}\sqrt{\frac{s_2p_2\log p}{2n}} + c_2d_{2,\mathrm{max}}\sqrt{\frac{s_1p_1\log p}{2n}} + c_1c_2\sqrt{s_1s_2}\frac{\log p}{2n}\\
        & \leq \sqrt{s_1s_2}\max \left[\frac{c_1d_{1,\mathrm{max}}}{\sqrt{s_1}}, \frac{c_2d_{2,\mathrm{max}}}{\sqrt{s_2}}\right](\sqrt{p_1}+\sqrt{p_2})\sqrt{\frac{\log p}{2n}} + c_1c_2\sqrt{s_1s_2}\frac{\log p}{2n}\\
        & \leq (1+\kappa)\sqrt{s_1s_2}\max \left[\frac{c_1d_{1,\mathrm{max}}}{\sqrt{s_1}}, \frac{c_2d_{2,\mathrm{max}}}{\sqrt{s_2}}\right](\sqrt{p_1}+\sqrt{p_2})\sqrt{\frac{\log p}{2n}}\\
        & \leq c\sqrt{\frac{(p_1+p_2)\log p}{n}}.
    \end{align}
    Here again for $\kappa$ the ratio factor
    \begin{equation}
         n \geq \frac{c_1^2c_2^2\log p}{2\kappa^2\max^2 \left[\frac{c_1d_{1,\mathrm{max}}}{\sqrt{s_1}}, \frac{c_2d_{2,\mathrm{max}}}{\sqrt{s_2}}\right]{(\sqrt{p_1}+\sqrt{p_2})}^2},
    \end{equation}
    and
    \begin{equation}
        c = (1+\kappa)\sqrt{s_1s_2}\max \left[\frac{c_1d_{1,\mathrm{max}}}{\sqrt{s_1}}, \frac{c_2d_{2,\mathrm{max}}}{\sqrt{s_2}}\right].
    \end{equation}
    Since the above also holds for ${\|\Delta_{\mathbf{W}^{(2t+1)}}\|}_F$, we have proved
    \begin{equation}
        {\|\Delta_{\mathbf{W}^{(t)}}\|}_F \leq c\sqrt{\frac{(p_1+p_2)\log p}{n}},\ \forall\  t\geq 2,
    \end{equation}
    with a high probability for sufficiently large $n$.
\end{proof}

\section{Competing Methods}
\label{sec:baselines}

\textbf{PST} is a GSP method that extracts the eigenvectors of factor GSOs from the signal covariance.
In \citep{einizade2023learning}, the GSO is set to be the weighted adjacency matrix of the product graph, which is the Kronecker product of the weighted adjacency matrices of the factor graphs $\mathbf{A}=\mathbf{A}_1\otimes\mathbf{A}_2$. 
In this case, the eigenvectors of the factor-wise covariance matrices converge to the eigenvectors of the factor-weighted adjacency matrices.
PST uses these eigenvectors, i.e. spectral templates, as a proxy, and solves for the eigenvalues that render a sparse graph.
Although this is generally not feasible when the GSO is Laplacian as $\mathbf{L}\neq\mathbf{L}_1\otimes\mathbf{L}_2$, it has been shown that for $\mathbf{L}=\mathbf{U}\mathbf{\Lambda}\mathbf{U}$, $\mathbf{L}_1=\mathbf{U}_1\mathbf{\Lambda}_1\mathbf{U}_1$, and $\mathbf{L}_2=\mathbf{U}_2\mathbf{\Lambda}_2\mathbf{U}_2$, the Kronecker product of factor Laplacian eigenvectors can approximate the eigenvectors of the product Laplacian $\mathbf{U}\approx\mathbf{U}_1\otimes\mathbf{U}_2$ \citep{sayama2016estimation,bavsic2022another}.
Let $\bm{\psi}=\textrm{vec}(\bm{\mathit{\Psi}})$ be the spectral representation of the smooth graph signals, we have
\begin{equation}
    \textrm{vec}(\bm{\mathit{X}}) = \bm{\mathit{x}} = \mathbf{U}\bm{\mathit{\phi}} \approx (\mathbf{U}_1\otimes\mathbf{U}_2)\bm{\mathit{\psi}} = \textrm{vec}(\mathbf{U}_2 \bm{\mathit{\Psi}} \mathbf{U}_1^T).
\end{equation}
Therefore the factor-wise covariance
\begin{align}
    & \mathbb{E}[\bm{\mathit{X}}\bm{\mathit{X}}^T] = \mathbb{E}[\mathbf{U}_2 \bm{\mathit{\Psi}} \mathbf{U}_1^T \mathbf{U}_1 \bm{\mathit{\Psi}}^T \mathbf{U}_2^T] = \mathbf{U}_2 \mathbb{E}[\bm{\mathit{\Psi}} \bm{\mathit{\Psi}}^T] \mathbf{U}_2^T,\\
    & \mathbb{E}[\bm{\mathit{X}}^T\bm{\mathit{X}}] = \mathbb{E}[\mathbf{U}_1 \bm{\mathit{\Psi}}^T \mathbf{U}_2^T \mathbf{U}_2 \bm{\mathit{\Psi}} \mathbf{U}_1^T] = \mathbf{U}_1^T \mathbb{E}[\bm{\mathit{\Psi}}^T \bm{\mathit{\Psi}}] \mathbf{U}_1.
\end{align}
Following Theorem 2 in \citep{einizade2023learning}, one can show that $\mathbb{E}[\bm{\mathit{\Psi}} \bm{\mathit{\Psi}}^T]$ and $\mathbb{E}[\bm{\mathit{\Psi}}^T \bm{\mathit{\Psi}}]$ are both diagonal matrices.
This means that the eigenvectors of the factor-wise covariance matrices approximate the eigenvectors of factor Laplacians.

\textbf{FF} is an alternating algorithm for estimating the Kronecker structured covariance matrices $\mathbf{\Sigma}=\mathbf{A}\otimes\mathbf{B}$ in matrix normal distributions.
By the property of the Kronecker product, $\det{(\mathbf{A}^{-1}\otimes\mathbf{B}^{-1}})=\det(\mathbf{A}^{-1})^{p_2}\det(\mathbf{B}^{-1})^{p_1}$.
Therefore, the MLE of the matrix normal distribution simplifies to a concise form
\begin{align}
    \min_{\mathbf{A} \in \mathbb{S}_{++}^{p_1},\mathbf{B} \in \mathbb{S}_{++}^{p_2}}  \left\{ \langle \mathbf{A}^{-1}\otimes\mathbf{B}^{-1}, \mathbf{S} \rangle - p_2\log{{\det}(\mathbf{A}^{-1})} - p_1\log{{\det}(\mathbf{B}^{-1})} \right\},
\end{align}
and FF alternates between closed-form updates of $\mathbf{A}$ and $\mathbf{B}$ to solve the problem.
\textbf{KGLasso} adds $\ell$-1 sparsity penalization to the problem and uses graphical lasso to solve each sub-problem.
Note that this MLE differs from our Kronecker graph learning problem, because the Laplacian of the Kronecker product graph cannot factor $\mathbf{L}\neq\mathbf{L}_1\otimes\mathbf{L}_2$.
This poses more difficulties in solving our penalized MLE as closed-form solutions no longer exist.
Therefore, using these GM methods to solve our problem causes a model mismatch.
\textbf{PGL} \citep{kadambari2021product}, \textbf{TeraLasso} \citep{greenewald2019tensor}, and \textbf{MWGL} \citep{shi2024learning} are Cartesian graph learning methods.
These methods are not appropriate for Kronecker product graph learning but serve as baselines for strong product graph learning.
Please refer to their original papers for the details.

\section{Additional Simulation Results}
\label{sec:pr-auc}

\begin{figure*}[hbt!]
\centering
\includegraphics[width=\textwidth]{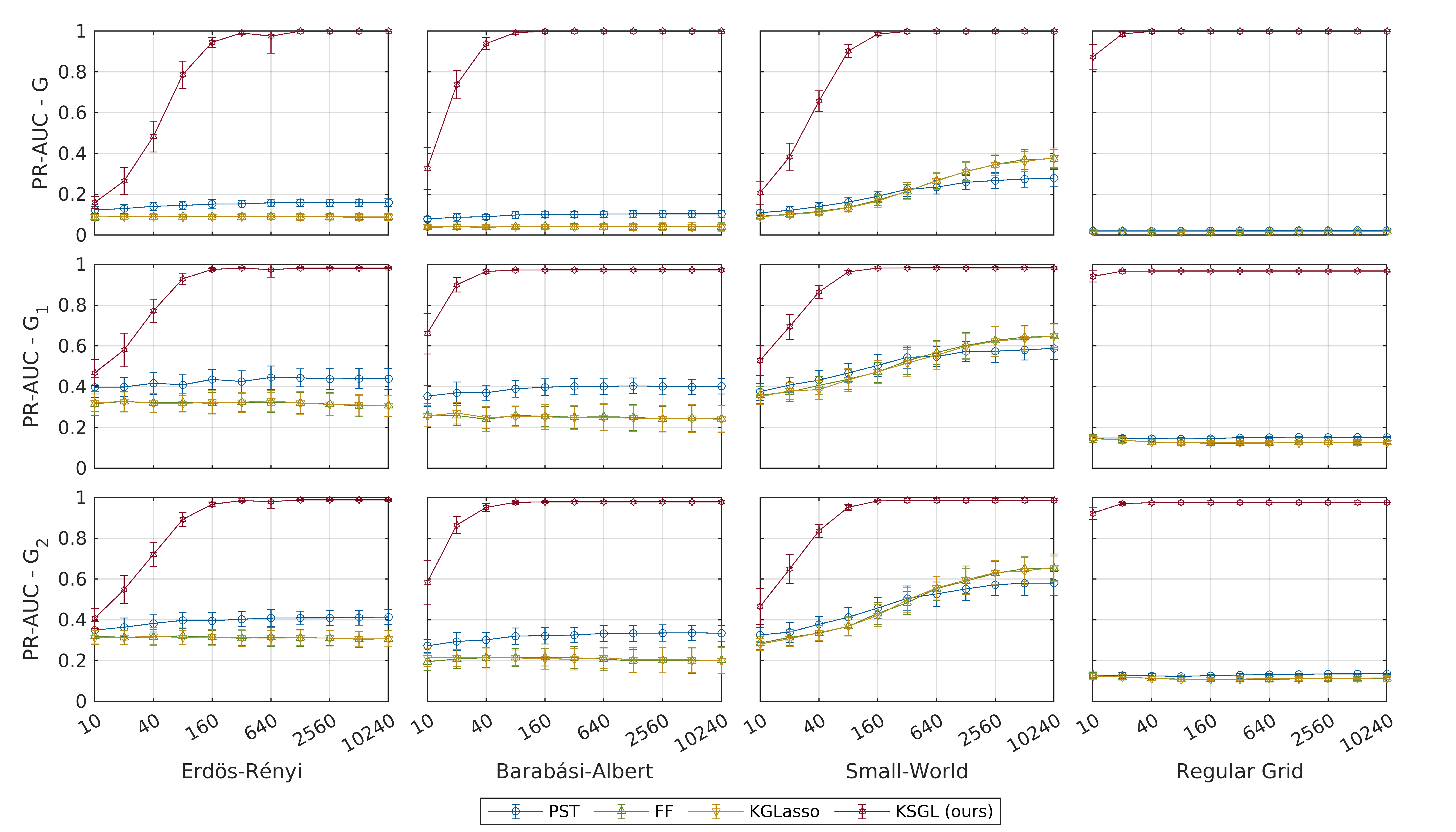}
\caption{Comparison of different methods on various synthetic Kronecker product graphs and signals.
Each sub-figure shows the trend of PR-AUC of the product or factor edge prediction as $n$ increases.}
\label{fig:prauc}
\end{figure*}

\begin{figure*}[hbt!]
\centering
\includegraphics[width=\textwidth]{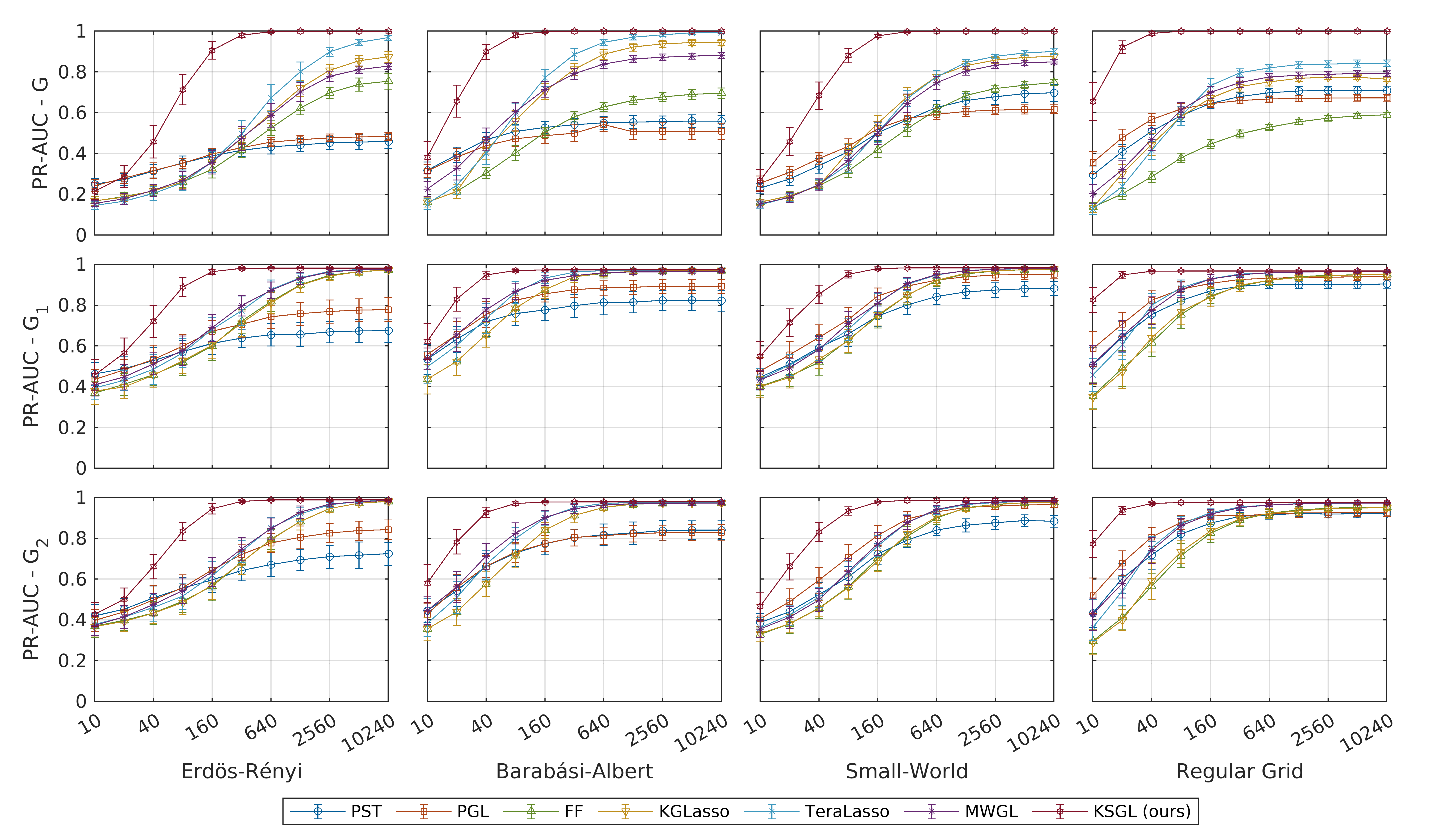}
\caption{Comparison of different methods on various synthetic strong product graphs and signals.
Each sub-figure shows the trend of PR-AUC of the product or factor edge prediction as $n$ increases.}
\label{fig:prauc_st}
\end{figure*}

\begin{figure*}[hbt!]
\centering
\includegraphics[width=0.7\textwidth]{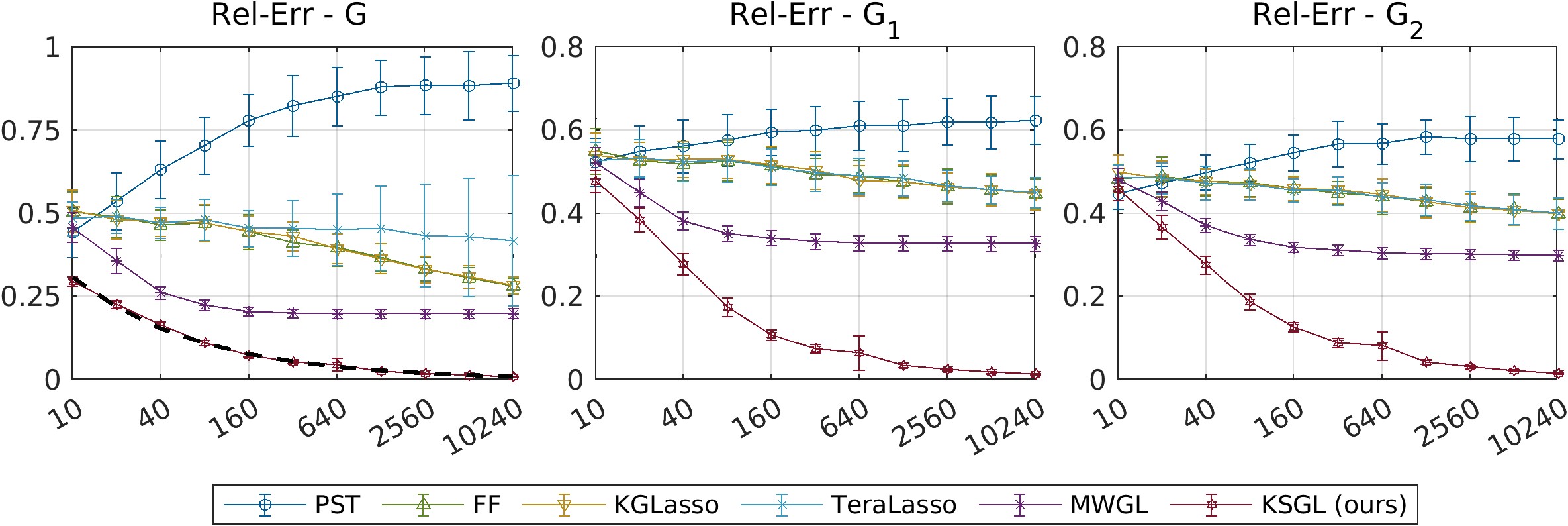}
\caption{Applying Cartesian product graph learning methods to learn Kronecker product graphs.}
\label{fig:rebuttal}
\end{figure*}

Here we provide additional results from synthetic experiments.
Fig.~\ref{fig:prauc} shows the PR-AUC of Kronecker product graph learning and Fig.~\ref{fig:prauc_st} shows the PR-AUC of strong product graph learning.
KSGL outperforms all competing methods on both Kronecker and strong product graph learning.
For Kronecker product graphs, only KSGL perfectly learns the true underlying graphs as the number of signals increases.
For strong product graphs, KSGL requires fewer graph signals to fully reconstruct the edge pattern.
Fig.~\ref{fig:rebuttal} shows the results of applying Cartesian product methods TeraLasso and MWGL to Kronecker product graph learning.
As we can see, these methods under-perform KSGL due to model mismatch as expected.

\section{Additional EEG Results}
\label{sec:addtional_eeg}

\begin{figure*}[hbt!]
    \centering
    \hspace{0.02\textwidth}
    \begin{subfigure}[b]{0.18\textwidth}
        \centering
        \includegraphics[width=\textwidth]{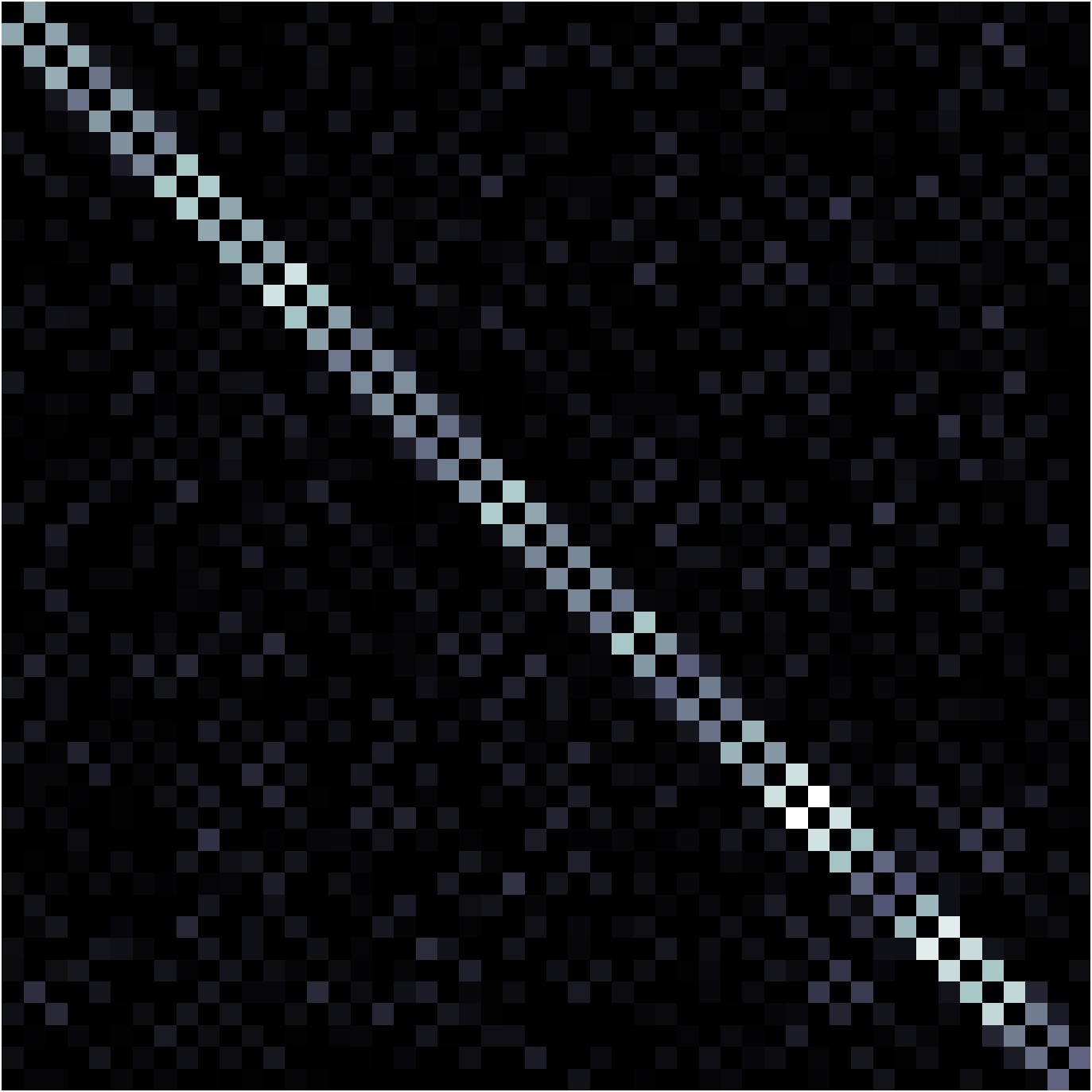}
        \caption{MWGL - normal}
    \end{subfigure}
    \hspace{0.02\textwidth}
    \begin{subfigure}[b]{0.18\textwidth}
        \centering
        \includegraphics[width=\textwidth]{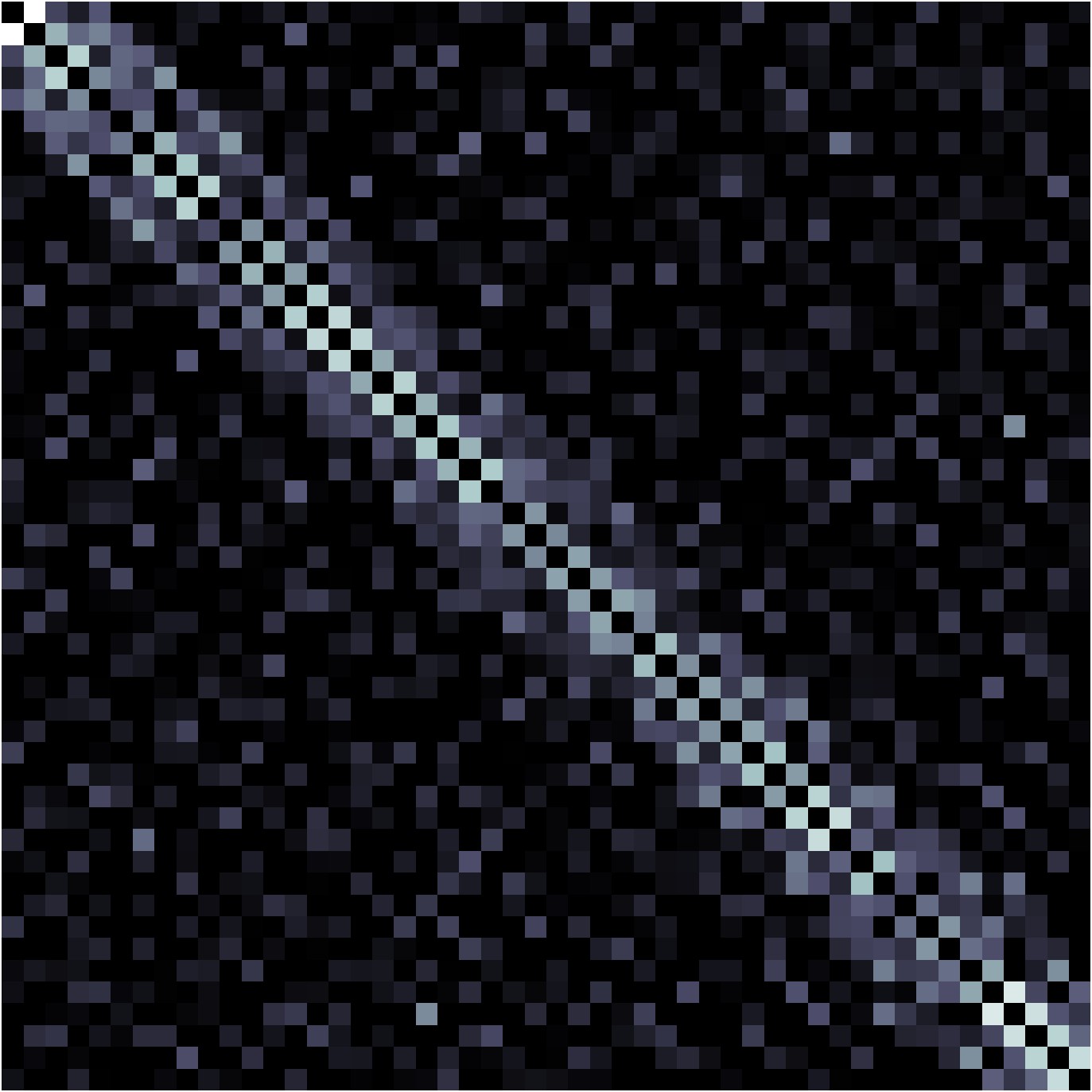}
        \caption{MWGL - seizure}
    \end{subfigure}
    \hspace{0.02\textwidth}
    \begin{subfigure}[b]{0.18\textwidth}
        \centering
        \includegraphics[width=\textwidth]{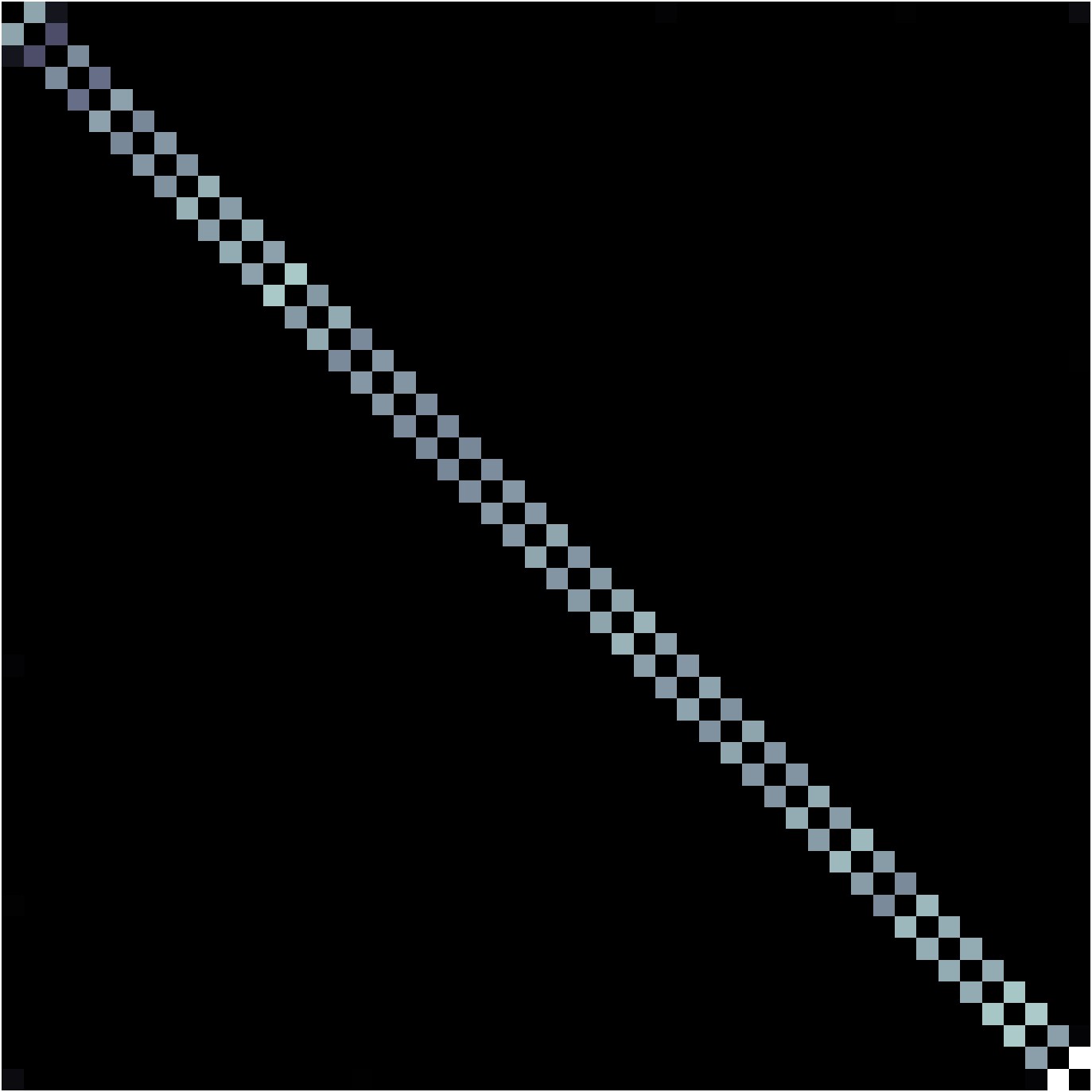}
        \caption{KSGL - normal}
    \end{subfigure}
    \hspace{0.02\textwidth}
    \begin{subfigure}[b]{0.18\textwidth}
        \centering
        \includegraphics[width=\textwidth]{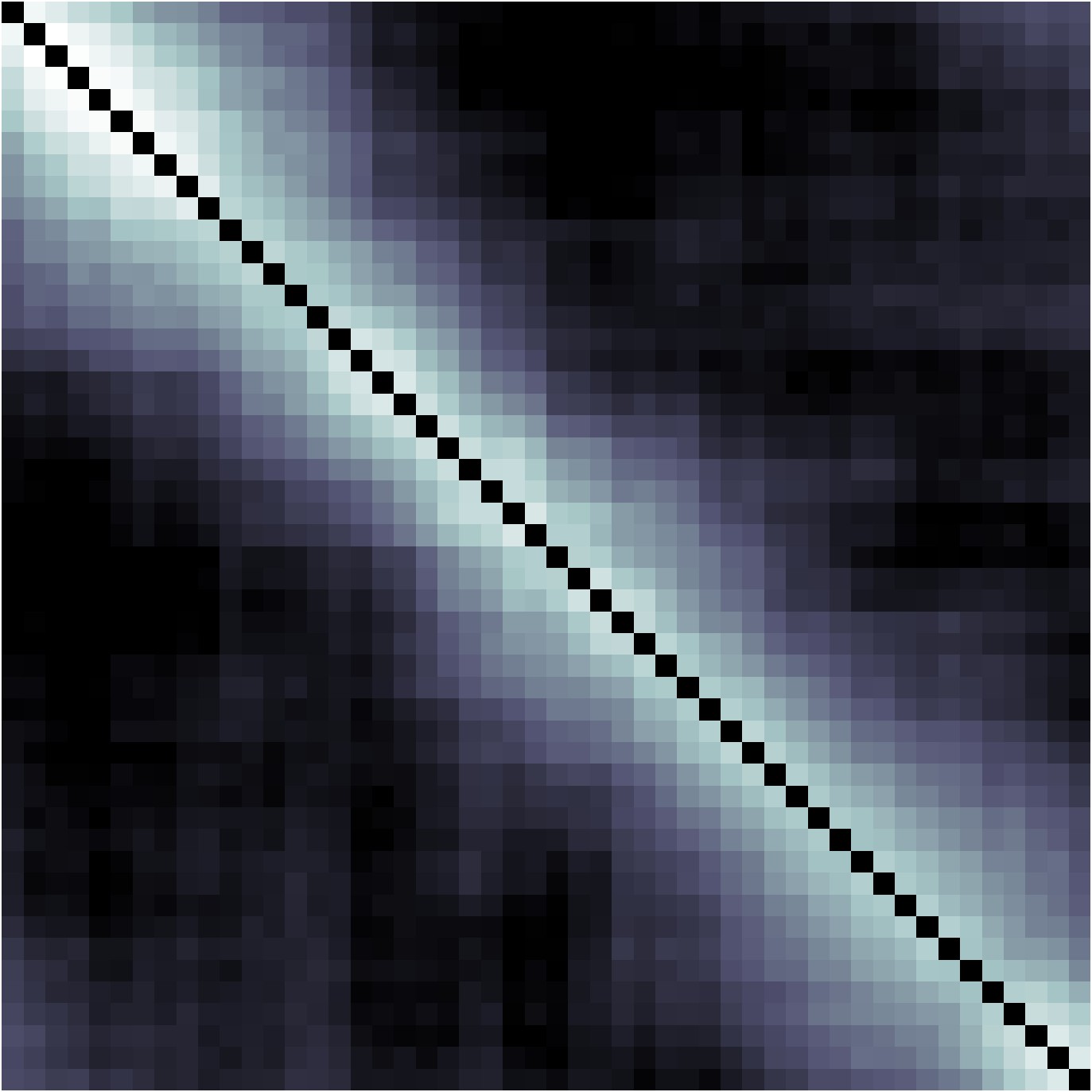}
        \caption{KSGL - seizure}
    \end{subfigure}
    \hspace{0.02\textwidth}
    \caption{The time adjacency matrices inferred by MWGL and KSGL. 
    }
    \label{fig:epileptic_time_graphs}
\end{figure*}

\begin{figure*}[hbt!]
\centering
\includegraphics[width=\textwidth]{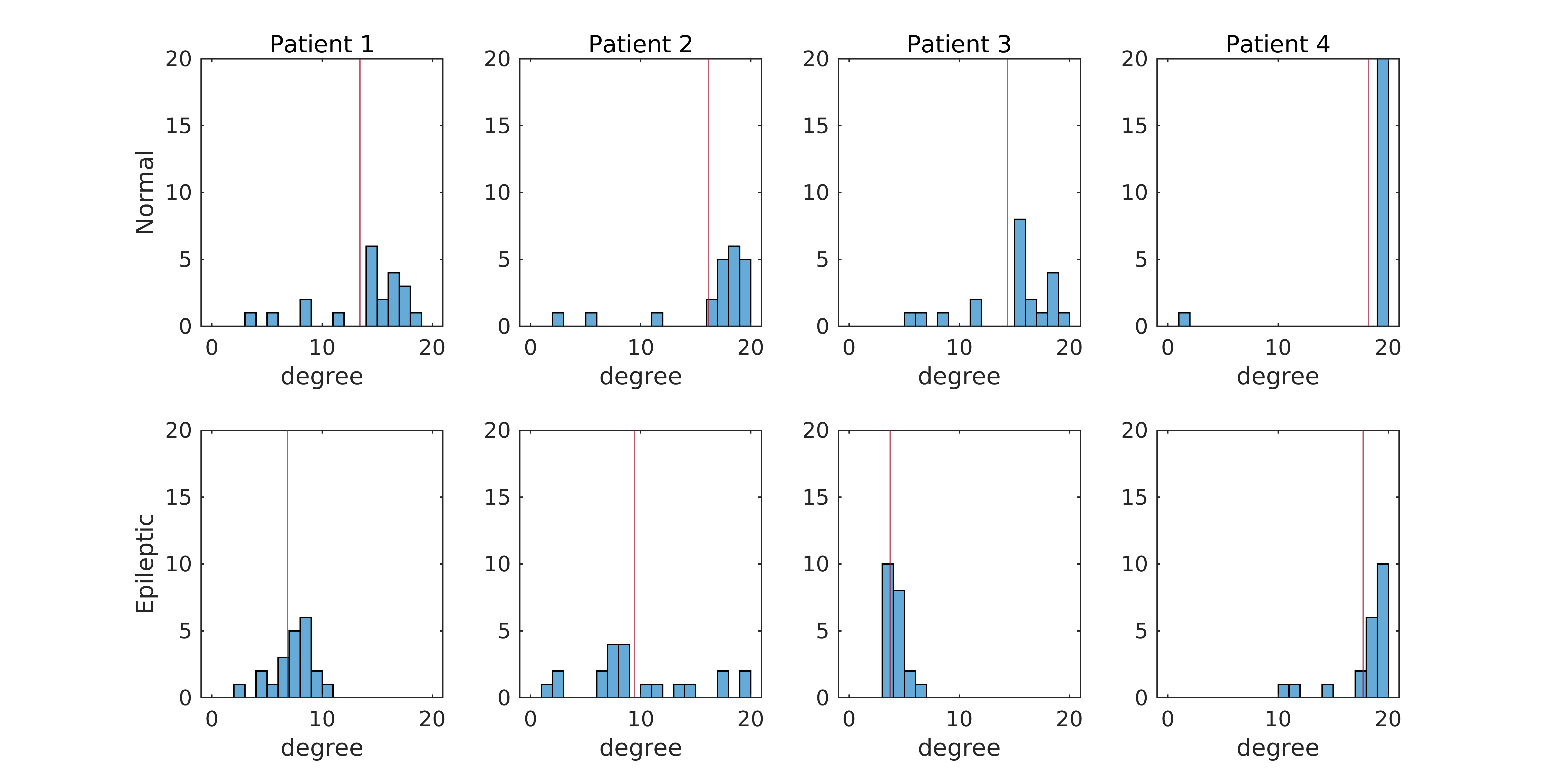}
\caption{Degree distributions of the learned brain graphs from normal and epileptic EEG of different patients.
The red vertical lines indicate the average node degree of the distributions.}
\label{fig:eeg_additional}
\end{figure*}

In Sec.~\ref{sec:exp_eeg}, we apply KSGL to the EEG signals of a patient whose seizures are not visible on EEG.
Although the signal amplitude does not manifest the ongoing seizures, the brain graph learned from epileptic signals shows increased connectivity compared with the normal one.
In Fig.~\ref{fig:epileptic_time_graphs}, we also show that KSGL learns more distinct temporal connectivity than MWGL, while the epileptic signals are more knitted than the normal signals for both methods.
Additionally, we select another type of patient whose seizures are visible on the EEG.
These patients all suffer from complex partial seizures, and we apply KSGL to their normal and epileptic EEG signals.
Fig.~\ref{fig:eeg_additional} shows the node degree distributions of the learned brain graphs.
We observe that KSGL learns denser connectivity from the normal EEG and sparser connectivity from the abnormal EEG, and this pattern is consistent across all four patients. 

\end{document}